\documentclass[runningheads]{llncs}
\usepackage[T1]{fontenc}

\usepackage{enumitem}
\usepackage{multirow}
\usepackage{graphicx}
\usepackage{diagbox}
\usepackage{footmisc} 

\usepackage[table,xcdraw]{xcolor}
\usepackage[dvipsnames]{xcolor}
\usepackage[colorlinks=true, linkcolor=blue, citecolor=green, urlcolor=magenta,backref]{hyperref}
\usepackage{cleveref}
\usepackage{booktabs}
\usepackage{multirow}
\usepackage{orcidlink}

\begin{document}
\newcommand{\msixdoc}{{\emph{M\textsuperscript{6}Doc}}}
\newcommand{\dfourla}{{D\textsuperscript{4}LA}}

\newcommand{\todo}[1]{\textcolor{black}{{#1}}}
\newcommand{\indicdlp}{{\emph{\textsc{IndicDLP}}}}

\newcommand{\rk}[1]{{\textcolor{green}{RK:  #1}}}
\newcommand{\mk}[1]{{\textcolor{red}{MK: #1}}}
\newcommand{\on}[1]{{\textcolor{blue}{ON:  #1}}}
\newcommand{\sk}[1]{{\textcolor{purple}{SK: #1}}}
\title{IndicDLP: A Foundational Dataset for Multi-Lingual and Multi-Domain Document Layout Parsing}
\titlerunning{IndicDLP}
%
\author{
Oikantik Nath\inst{1}\textsuperscript{[0009-0005-5407-0455]} \and
Sahithi Kukkala\inst{2}\textsuperscript{[0009-0003-5483-2232]} \and \\
Mitesh Khapra\inst{1}\textsuperscript{[0009-0008-3687-9922]} \and \\
Ravi Kiran Sarvadevabhatla\inst{2}\textsuperscript{[0000-0003-4134-1154]}
}
\authorrunning{Nath et al.}
%
\institute{Indian Institute of Technology, Madras, India\\
\email{\{oikantik,miteshk\}@cse.iitm.ac.in}\\ \and
International Institute of Information Technology Hyderabad, India\\
\email{\{ravi.kiran@,sahithi.kukkala@research.\}iiit.ac.in}\\
}

\maketitle              
\begingroup\renewcommand\thefootnote{}\footnote{\texttt{Project Page:} \url{https://indicdlp.github.io/}}\endgroup

\begin{abstract}

Document layout analysis is essential for downstream tasks such as information retrieval, extraction, OCR, and digitization. However, existing large-scale datasets like PubLayNet and DocBank lack fine-grained region labels and multilingual diversity, making them insufficient for representing complex document layouts. Human-annotated datasets such as \msixdoc~and \dfourla~offer richer labels and greater domain diversity, but are too small to train robust models and lack adequate multilingual coverage. This gap is especially pronounced for Indic documents, which encompass diverse scripts yet remain underrepresented in current datasets, further limiting progress in this space. To address these shortcomings, we introduce \indicdlp, a large-scale foundational document layout dataset spanning 11 representative Indic languages alongside English and 12 common document domains. Additionally, we curate \textsc{UED-mini}, a dataset derived from DocLayNet and \msixdoc, to enhance pretraining and provide a solid foundation for Indic layout models. Our experiments demonstrate that fine-tuning existing English models on \indicdlp~significantly boosts performance, validating its effectiveness. Moreover, models trained on \indicdlp~generalize well beyond Indic layouts, making it a valuable resource for document digitization. This work bridges gaps in scale, diversity, and annotation granularity, driving inclusive and efficient document understanding.
\end{abstract}


\section{Introduction}

Document Layout Parsing (DLP), also known as Document Layout Segmentation, is a fundamental task in document understanding that enables downstream applications such as information extraction~\cite{Ha2022}, retrieval~\cite{Jaha2024}, OCR~\cite{Fateh2024}, and automated document conversion~\cite{Auer2022}. 
 However, DLP remains highly challenging due to the diverse layouts, font styles, and structural variations found across different document types.
Given its importance, several datasets have been introduced in recent years, either automatically derived from existing digital document metadata~\cite{publaynet,docbank} or manually labeled~\cite{m6doc,dfourla,doclaynet}. While automatically generated datasets offer large-scale training data, they often lack fine-grained annotations. In contrast, manually labeled datasets provide higher-quality annotations, but are limited in scale. A common limitation across both types is the lack of multilingual coverage, with most datasets supporting only English and a few resource-rich languages. This resource gap has significantly hindered progress for low-resource languages, especially those from the Indian subcontinent, which feature diverse scripts distinct from Latin-based ones. As our results shall demonstrate (\Cref{fig:comparisons-qualitative}), models trained on English documents have subpar generalization to Indian languages.

To bridge this gap, we introduce \indicdlp, the largest and most diverse Indic document layout dataset to date, consisting of 119,806 human-annotated images across 12 domains, 11 Indic languages alongside English, and 42 region labels. The dataset enables robust training for layout parsing across a wide range of document types, including newspapers, magazines, novels, textbooks, acts \& rules, notices, manuals, syllabi, question papers, forms, brochures, and research papers. To ensure a comprehensive yet minimal label set, we aligned our annotation schema with the \msixdoc~\cite{m6doc} guidelines, refining label granularity through manual analysis of \indicdlp~images. The annotation process followed a maker-checker workflow with over 50 annotators and reviewers per language, guided by a detailed 150-page manual. Over 60\% of the annotations were further validated for cross-domain consistency by a team of 8 supercheckers.

\begin{figure}[!t]
\centering
\includegraphics[width=0.80\linewidth]{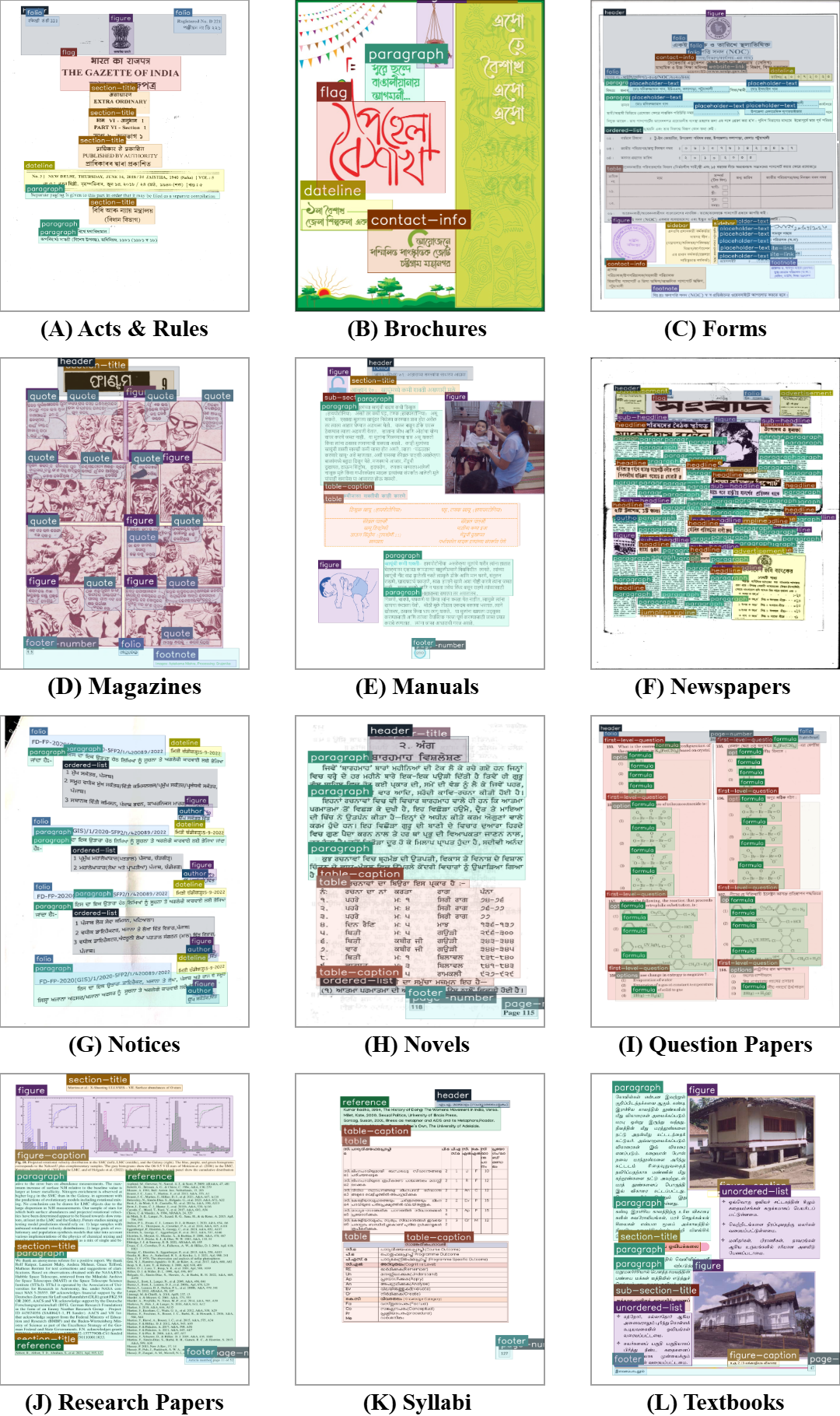}
\caption{Samples from the \indicdlp~dataset highlighting its diversity across document formats, domains, languages, and temporal span.
For improved differentiability, segmentation masks are used instead of bounding boxes to highlight regions more effectively. 
This figure is best viewed when zoomed in.
}
\label{fig:dataset}
\end{figure}

Using \indicdlp, we conduct a series of experiments to evaluate the performance of existing object detection and layout parsing models on Indic document layout parsing. We establish baseline performance by fine-tuning state-of-the-art models on \indicdlp, including YOLOv10~\cite{yolov10}, DiT~\cite{dit}, DocLayout-YOLO~\cite{doclayout-yolo}, DINO~\cite{dino}, RoDLA~\cite{rodla}, and a vision language model Florence-2~\cite{florence2}. Our results show significant performance variations across domains, with models struggling on documents with more unique regions and complex multicolumn layouts. 
Performance also varies substantially across languages, with zero-shot evaluation on unseen languages showing a 20–25 mAP point drop compared to those seen during training, suggesting that script-based variations influence layout parsing performance. 

Additionally, we explore the utility of \indicdlp~as a pretraining resource for document layout parsing, observing further performance gains and faster convergence. To facilitate future research on language-specific document understanding, we release \indicdlp~along with trained models, datasets, and evaluation scripts.

\section{Related Work}

\paragraph{Layout Parsing Datasets.}

The inherent challenges of deep learning-based document parsing have prompted the development of various datasets, each with distinct limitations. Large-scale datasets like PubLayNet~\cite{publaynet} and DocBank~\cite{docbank} offer ample size but are constrained by limited region labels and domain diversity. The RVL-CDIP dataset~\cite{rvlcdip} improves domain diversity for document classification tasks but uses grayscale images, restricting its relevance for modern layouts. Similarly, \dfourla~\cite{dfourla}, a manually annotated subset of RVL-CDIP designed for layout parsing, faces the same limitation. Datasets such as \msixdoc~\cite{m6doc}, DocLayNet~\cite{doclaynet}, and BaDLAD~\cite{badlad} mitigate some of these issues by incorporating diverse domains and document layouts, but still lack source variety and suffer from overly fine-grained labels. Moreover, \msixdoc~\cite{m6doc} reports poor cross-dataset performance between \msixdoc~ and DocLayNet~\cite{doclaynet}, highlighting the need for a more diverse dataset that covers a broad range of domains and provides an adequate representation of common document types. As we show in \Cref{indicdlp_desc}, our dataset \indicdlp~addresses these shortcomings, offering broader coverage and better representation of document types.






\begin{table*}[!t]
\renewcommand{\arraystretch}{1.1}
\centering
\scriptsize
\resizebox{\textwidth}{!}{%
\begin{tabular}{lrrrrr}
\toprule
\textbf{Dataset} & \textbf{\#Images} & \textbf{\#Region Classes} & \textbf{Annotation Method} & \textbf{\#Domains} & \textbf{\#Languages} \\ 
\midrule
PRImA~\cite{prima}       & 1,240     & 10  & Automatic       & 5  & 1  \\
PubLayNet~\cite{publaynet}  & 360,000  & 5   & Automatic       & 1  & 1  \\
DocBank~\cite{docbank}    & 500,000  & 13  & Automatic       & 1  & 1  \\
DocLayNet~\cite{doclaynet}  & 80,863   & 11  & \textbf{Manual} & 6  & 4  \\
\msixdoc~\cite{m6doc}     & 9,080    & \textbf{75}  & \textbf{Manual} & 7  & 2  \\
D$^4$LA~\cite{dfourla}    & 11,092   & 27  & \textbf{Manual} & 12  & 1  \\
BaDLAD~\cite{badlad}      & 33,695   & 4   & \textbf{Manual} & 6  & 1  \\
\textbf{\indicdlp (Ours)}  & \textbf{\todo{121,198}} & \todo{42} & \textbf{Manual} & \textbf{12} & \textbf{12} \\ 
\bottomrule
\end{tabular}%
}
\vspace{5pt} 
\caption{Comparison of modern document layout parsing datasets.}
\label{tab:datasets}
\end{table*}

\paragraph{Layout Parsing Models.}

Existing research on layout parsing has primarily focused on visual approaches, advancing document understanding through models like RCNNs~\cite{fasterrcnn,maskrcnn,cascadercnn} and single-shot detectors such as SSD~\cite{ssd} and the YOLO series~\cite{yolo-survey}. Transformers further improved layout parsing performance with models like DocSegTr~\cite{docsegtr}, SwinDocSegmenter~\cite{swindocsegmenter}, and TransDLANet~\cite{m6doc}. While DiT~\cite{dit} leveraged self-supervised pretraining, DocLayout-YOLO~\cite{doclayout-yolo} used synthetic pretraining for competitive results. RoDLA~\cite{rodla} introduced robustness enhancements via a variant of DINO~\cite{dino}. More recently, multimodal models incorporating textual, visual, and spatial features~\cite{m2doc,docformer,udop,selfdoc} have emerged, but their effectiveness in modality integration is limited by the scarcity of multilingual textual data, especially for Indian languages with limited OCR resources. Therefore, this work focuses on visual document layout parsing models to highlight state-of-the-art performance on multidomain Indic document data.

\section{\indicdlp~}
\label{indicdlp_desc}
\subsection{Motivation}

To build useful and general layout detection models, a document layout dataset should (i) be large in terms of document count (ii) span a diverse and representative set of domain categories (e.g. newspapers, forms, textbooks) (iii) provide a comprehensive set of region labels with sufficient label frequency (iv) cover a large set of languages. For the remainder of this section, we describe our dataset and demonstrate how it meets the above criteria.

At first glance, the language coverage criteria do not seem very obvious since layout detection appears to be script agnostic. However, many Indic languages have intricate scripts characterized by inflections, diacritics, and composite or conjugated characters which complicates layout detection (see \Cref{fig:dataset}). Also, a key feature of many Indic scripts is that the visual form of certain characters in a word changes based on their interaction with neighboring letters. These script-level complexities are further amplified by the prevalence of scanned copies or photographs as the primary form of publicly available Indic documents, limiting the effectiveness of synthetic methods for automatic annotation. In addition, many printed Indic documents span multiple decades in origin and exhibit significant typographical and printing variations. These historical texts pose challenges due to their complex layouts, frequently illegible writing, deteriorated or stained paper quality, and non-standard formatting, making automated processing especially difficult. 

From a downstream application perspective, poor localization negatively affects the performance of transcription pipelines for Optical Character Recognition (OCR). To address this, a high-quality and robust general layout parsing model trained on manually-annotated diverse Indic documents, spanning multiple scripts, domains and formats, is required.

\begin{table*}[!t]
\renewcommand{\arraystretch}{1.1}
\scriptsize
\centering
\begin{tabular*}{\textwidth}{@{\extracolsep{\fill}}lrrrr@{}}
\toprule
\textbf{Domain} & \textbf{\#Inst} & \textbf{\%} & \textbf{ARD} & \textbf{AAT(s)} \\ 
\midrule
Acts \& Rules \textbf{(ar)} & 12,210 & 10.074 & 10 & 766 \\
Brochures \textbf{(br}) & 10,128 & 8.357 & 14 & 1207 \\
Forms \textbf{(fm)} & 6,751 & 5.570 & 23 & 1196 \\
Magazines \textbf{(mg}) & 12,807 & 10.567 & 15 & 1667 \\
Manuals \textbf{(mn)} & 11,660 & 9.621 & 10 & 330 \\
Newspapers \textbf{(np)} & 6,886 & 5.682 & 72 & 3682 \\
Notices \textbf{(nt)} & 6,602 & 5.447 & 11 & 354 \\
Novels \textbf{(nv)} & 13,385 & 11.044 & 10 & 304 \\
Question Papers \textbf{(qp)} & 7,866 & 6.490 & 17 & 708 \\
Research Papers (\textbf{rp)} & 11,435 & 9.435 & 9 & 322 \\
Syllabi \textbf{(sy)} & 8,222 & 6.784 & 11 & 445 \\
Textbooks \textbf{(tb)} & 13,246 & 10.929 & 11 & 520 \\
\midrule
\textbf{Total} & \textbf{121,198} & \textbf{100} & {} & {} \\
\bottomrule
\end{tabular*} \\
\vspace{5pt} 

\caption{Overview of the different domains in \indicdlp~, including the total number of documents, percentage contribution to \indicdlp, average number of regions per document [ARD] (rounded to the nearest integer), and average annotation time per document [AAT] (in seconds). The disproportionately high average annotation time for newspapers is due to the large number of regions they typically contain.}
\label{tab:document_category_distribution}
\end{table*}
\subsection{Dataset domains and sources}

\indicdlp~comprises \todo{119,809} layout-annotated document images in 11 different Indian languages and English. These documents span \todo{12} document domain categories: newspapers (\todo{5.3\%}), novels (\todo{11\%}), magazines (\todo{10.6\%}), manuals (\todo{9.6\%}), forms (\todo{5.5\%}), textbooks (\todo{10.9\%}), acts and rules (\todo{10.1\%}), question papers (\todo{6.5\%}), brochures (\todo{8.3\%}), notices (\todo{5.4\%}) and syllabi (\todo{6.7\%}), and research papers (\todo{9.5\%}). The dataset covers a wide range of formats, including documents from the pre-independence era to the present day.

The document images are sourced from various digital-born files, scanned documents, or photographed pages. In total, they include \todo{1,856,241} annotated instances, encompassing \todo{42} physical and logical region labels. In the following sections, we detail our data curation process (\Cref{domain_categories}), our methodology for selecting region labels (\Cref{region_label_set}) and the annotation and review workflow (\Cref{annot_review_workflow}).

\subsection{Domain Categories}
\label{domain_categories}

To ensure dataset diversity, we curated documents encountered in popular use case scenarios. The sources for each category are detailed below, and representative samples across different languages are shown in \Cref{fig:dataset}.

\subsubsection{Newspapers:}

This domain includes both modern and historical newspapers. Modern ones are digitally generated or high-quality scans, while historical scans exhibit older fonts, disjoint lettering, and unique glyphs like conjuncts. These documents often feature multi-page, multi-column, multi-section layouts with diverse labels, making them among the most complex domains in the dataset.

\noindent \textbf{Magazines \& Manuals:} Magazines and manuals typically contain single or double-page multi-column layouts. We include image-rich comics and advertisement magazines from online sources to enhance layout diversity.

\noindent \textbf{Novels:} Indic novels offer insights into traditional typesetting and typography. We curated PDFs from various non-copyrighted ebook stores, including covers, tables of contents, and indices wherever available, to increase layout variety.

\noindent \textbf{Textbooks:} Textbooks used in Indian schools with native language teaching mediums are typically written in regional languages. They often feature interleaved text and images, contributing to visual and structural diversity.

\noindent \textbf{Acts \& Rules:} We include digitally generated or scanned public documents and bulletins released by the Indian Government. These documents feature logos, seals, and intricate headers, enriching the variety of layouts.

\noindent \textbf{Question Papers:} Indian schools and national exams often provide bilingual question papers in English or Hindi along with a native language. These documents feature rich multi-script regions within a single source, making them valuable for training models to handle script diversity effectively.

\noindent \textbf{Forms:} Forms feature structured fields, tables, and handwritten input areas. Our collection includes both government-issued and organizational forms in both digital and scanned formats.

\noindent \textbf{Brochures:} Brochures, pamphlets, and leaflets typically feature diverse graphical layouts with sparse text. \indicdlp~includes brochures from government agencies, educational institutions, and various commercial sources such as retail, healthcare, tourism, finance, and real estate, capturing diverse design styles.

\noindent \textbf{Notices \& Syllabi:} Structured with headers, lists, and tables, these documents come from schools, universities, online bulletin boards and government sources, offering a wide range of academic and administrative layouts.

\noindent \textbf{Research Papers:} Indian research papers, theses, and multidisciplinary scholarly articles are a rich source of structured academic writing, typically formatted in single or double-column layouts with tables, figures, and citations. Our dataset includes research articles downloaded as PDFs from open-access repositories such as Shodhganga \cite{shodhganga} and arXiv, capturing diverse academic styles, equations, and multilingual content.

\subsection{Region Label Set Curation}
\label{region_label_set}

\begin{table*}[!t]
\renewcommand{\arraystretch}{1.1}
\scriptsize
\centering
\begin{tabular*}{\textwidth}{@{\extracolsep{\fill}}lrrrrrr@{}}
\toprule
\multicolumn{1}{l}{\multirow{2}{*}{\textbf{Region Labels}}} & \multicolumn{2}{c}{\textbf{Train}} & \multicolumn{2}{c}{\textbf{Val}} & \multicolumn{2}{c}{\textbf{Test}} \\
\multicolumn{1}{l}{} & \multicolumn{1}{r}{\textbf{\#Inst}} & \multicolumn{1}{c}{\textbf{\%}} & \multicolumn{1}{r}{\textbf{\#Inst}} & \multicolumn{1}{c}{\textbf{\%}} & \multicolumn{1}{r}{\textbf{\#Inst}} & \multicolumn{1}{c}{\textbf{\%}} \\ 
\midrule
advertisement & 19175 & 1.293 & 2038 & 1.086 & 2110 & 1.134 \\
answer & 4100 & 0.277 & 449 & 0.239 & 556 & 0.299 \\
author & 17860 & 1.205 & 2247 & 1.197 & 2171 & 1.167 \\
chapter-title & 6471 & 0.436 & 833 & 0.444 & 781 & 0.420 \\
contact-info & 12070 & 0.814 & 1695 & 0.903 & 1538 & 0.827 \\
dateline & 31158 & 2.102 & 3853 & 2.053 & 3925 & 2.110 \\
figure & 76957 & 5.191 & 9735 & 5.187 & 10022 & 5.388 \\
figure-caption & 25202 & 1.700 & 3201 & 1.706 & 3396 & 1.826 \\
first-level-question & 37009 & 2.496 & 4742 & 2.527 & 4838 & 2.601 \\
flag & 3337 & 0.225 & 424 & 0.226 & 447 & 0.240 \\
folio & 82112 & 5.538 & 10501 & 5.595 & 10476 & 5.632 \\
footer & 55787 & 3.763 & 7068 & 3.766 & 6974 & 3.749 \\
footnote & 12304 & 0.830 & 1557 & 0.830 & 1539 & 0.827 \\
formula & 15157 & 1.022 & 1959 & 1.044 & 1883 & 1.012 \\
header & 62207 & 4.196 & 7867 & 4.192 & 7842 & 4.216 \\
headline & 44758 & 3.019 & 5557 & 2.961 & 5480 & 2.946 \\
index & 460 & 0.031 & 81 & 0.043 & 33 & 0.018 \\
jumpline & 3897 & 0.263 & 544 & 0.290 & 512 & 0.275 \\
options & 35575 & 2.400 & 4518 & 2.407 & 4699 & 2.526 \\
ordered-list & 40672 & 2.743 & 5232 & 2.788 & 5017 & 2.697 \\
page-number & 79316 & 5.350 & 10025 & 5.342 & 9933 & 5.340 \\
paragraph & 479891 & 32.369 & 60964 & 32.485 & 59860 & 32.183 \\
placeholder-text & 77291 & 5.213 & 9367 & 4.991 & 9708 & 5.219 \\
quote & 3495 & 0.236 & 455 & 0.242 & 337 & 0.181 \\
reference & 5442 & 0.367 & 700 & 0.373 & 658 & 0.354 \\
second-level-question & 13203 & 0.891 & 1725 & 0.919 & 1572 & 0.845 \\
section-title & 81270 & 5.482 & 10542 & 5.617 & 10123 & 5.443 \\
sidebar & 39868 & 2.689 & 5182 & 2.761 & 4847 & 2.606 \\
sub-headline & 15226 & 1.027 & 1928 & 1.027 & 2105 & 1.132 \\
sub-ordered-list & 18795 & 1.268 & 2399 & 1.278 & 2313 & 1.244 \\
sub-section-title & 17908 & 1.208 & 2164 & 1.153 & 2333 & 1.254 \\
sub-unordered-list & 866 & 0.058 & 90 & 0.048 & 70 & 0.038 \\
subsub-headline & 1892 & 0.128 & 201 & 0.107 & 235 & 0.126 \\
subsub-ordered-list & 4455 & 0.300 & 610 & 0.325 & 555 & 0.298 \\
subsub-section-title & 2506 & 0.169 & 237 & 0.126 & 274 & 0.147 \\
subsub-unordered-list & 96 & 0.006 & 3 & 0.002 & 8 & 0.004 \\
table & 19389 & 1.308 & 2407 & 1.283 & 2402 & 1.291 \\
table-caption & 7376 & 0.498 & 920 & 0.490 & 930 & 0.500 \\
table-of-contents & 956 & 0.064 & 135 & 0.072 & 109 & 0.059 \\
third-level-question & 2669 & 0.180 & 265 & 0.141 & 252 & 0.135 \\
unordered-list & 12795 & 0.863 & 1652 & 0.880 & 1676 & 0.901 \\
website-link & 11599 & 0.782 & 1598 & 0.851 & 1460 & 0.785 \\
\midrule
\textbf{Total} & \textbf{1482572} & 100 & \textbf{187670} & 100 & \textbf{185999} & 100 \\
\bottomrule
\end{tabular*}
\vspace{5pt} 
\caption{Distribution of layout regions in \indicdlp, sorted in alphabetical order of region label names.}
\label{tab:dataset_overview}
\end{table*}

We analyzed existing layout parsing datasets to define a concise yet comprehensive label set for document understanding tasks. To ensure broad coverage, we followed \msixdoc~\cite{m6doc} guidelines, incorporating region labels for specific document categories. This was crucial for domains like newspapers and forms, which contain unique elements such as \textit{jumplines} and \textit{placeholders}, elements typically absent from other domains.

To establish an initial set of region labels covering over 90\% of common document structures, we also reviewed several other layout parsing datasets. Consistent with \msixdoc~\cite{m6doc} findings, we observed that datasets like PubLayNet~\cite{publaynet}, DocBank~\cite{docbank}, and DocLayNet~\cite{doclaynet} primarily focus on basic, domain-independent labels. To achieve finer granularity, we expanded the label set by an additional \todo{31} categories. For example, we split \textit{list} into \textit{ordered} and \textit{unordered}, and \textit{caption} into \textit{table caption} and \textit{figure caption}. Our dataset also includes multi-level hierarchical labels for elements such as \textit{headlines}, \textit{section titles}, and \textit{lists}.

Further refinement was based on a careful manual analysis of our dataset images, focusing on the frequency and visual prominence of specific regions. We excluded rarely occurring labels like \textit{QR code}, \textit{bar code}, and \textit{poem} (included in \msixdoc~\cite{m6doc}) to reduce complexity. Instead, we included frequently encountered regions such as \textit{website links}, \textit{contact information}, and \textit{quotations}, which are common in magazines and newspapers. The complete set of region labels and their instance counts are presented in \Cref{tab:dataset_overview}.

\subsection{Annotation and Review Workflow}
\label{annot_review_workflow}

Our primary objective in this work is to develop a uniform and consistent set of annotated document pages for layout parsing across multiple languages. To achieve this, we trained a team of 50 individuals, comprising 3 to 4 annotators and 1 reviewer per language. We developed a 150-page guideline to ensure clear and consistent region-based annotation, supplemented with annotated examples spanning multiple languages and domains. Rectangular bounding boxes were chosen for their consistency across languages and domains, enabling efficient large-scale annotation. This choice offered a practical trade-off between the higher precision of polygonal or rotated boxes in skewed cases and overall annotation efficiency. We point out several key annotation decisions from the guidelines that are different from datasets like \msixdoc~\cite{m6doc}, PubLayNet~\cite{publaynet} and DocLayNet~\cite{doclaynet}:

\begin{itemize}[label=$\bullet$] 
    \item Figures include plots, graphs, barcodes, seals, QR codes, and logos.
    \item Text within figures and sub-figures is \textit{not} annotated separately.
    \item Variable names in research papers are \textit{not} labeled as formulae to reduce ambiguity, unlike \msixdoc~\cite{m6doc}. Clear guidelines specify when mathematical notation should be annotated as formulae.
    \item Items in the header or footer sections are labeled appropriately and enclosed within a \textit{header} or \textit{footer} region container, respectively.
\end{itemize}

All annotations were performed using Shoonya \cite{shoonya}, an open-source annotation framework based on Label Studio. We meticulously tracked metadata, including annotation performance, average annotation time, draft mode usage, and correction records. Each annotation was validated by a language-proficient reviewer, and a team of 10 supercheckers, including the authors, further checked over 60\% of the validated documents to ensure cross-language and cross-domain consistency. The entire annotation process, including two-step validation, was completed in approximately eight months. Finally, for dataset preparation, images were scaled to a maximum of 1024 pixels on the shortest side and 1333 pixels on the longest. The dataset was split into training, validation, and testing sets in an 80:10:10 ratio using a stratified approach to preserve language and domain proportions. The distribution of region labels for each split is presented in \Cref{tab:dataset_overview}. Further details on language-wise annotated data distribution and annotation workflow can be found in our project page.

\section{Experiments}

In this section, we present a comprehensive qualitative and quantitative evaluation of various state-of-the-art (SOTA) document layout parsing and object detection models trained on the \indicdlp~dataset. Object detection is chosen as the primary task for layout parsing, with mean Average Precision (mAP@[.5:.95]) as the performance metric, calculated over thresholds ranging from 0.5 to 0.95 in steps of 0.05. We trained several state-of-the-art models, including YOLOv10~\cite{yolov10}, DiT~\cite{dit}, DocLayout-YOLO~\cite{doclayout-yolo}, DINO~\cite{dino}, RoDLA~\cite{rodla}, and Florence-2~\cite{florence2}, exploring different model sizes and architectures to establish baselines. Original hyperparameters, as specified in the respective papers, were retained for our experiments. All models were trained and evaluated exclusively on our \indicdlp~dataset on 8 NVIDIA H100 GPUs. Detailed quantitative results are provided in Table~\ref{tab:performance_comparison}.

\subsection{How do different models perform on \indicdlp~?}
As shown in \Cref{tab:performance_comparison}, the diverse region labels and their varying appearances across domains in \indicdlp~pose significant challenges for all models.
 Notably, DiT~\cite{dit}, RoDLA~\cite{rodla}, and DocLayout-YOLO~\cite{doclayout-yolo} were the only models pretrained on document-based datasets: IIT-CDIP, \msixdoc~\cite{m6doc}, and DocSynth-300K \cite{doclayout-yolo}, respectively. Except for Florence-2~\cite{florence2}, which was pretrained on the FLD-5B dataset, all other models were pretrained on the COCO natural images dataset~\cite{coco}. Among them, YOLOv10x~\cite{yolov10} emerges as one of the top performers for layout parsing. Therefore, we conduct further ablation and fine-tuning analysis on YOLOv10x~\cite{yolov10} due to its high efficiency and low memory footprint.

\begin{table*}[!t]
\renewcommand{\arraystretch}{1.1}
\scriptsize
\centering
\begin{tabular}{@{}lcccc@{}}
\toprule
\textbf{Model}                & \textbf{Model Size(M)}  & \textbf{mAP50($\uparrow$)}  & \textbf{mAP75($\uparrow$)} & \textbf{mAP[50-95]($\uparrow$)} \\ 
\midrule
\textbf{YOLOv10x \cite{yolov10}} & \textbf{37} & 73.5 & \textbf{60.6} & \textbf{55.0} \\ 
DocLayout-YOLO \cite{doclayout-yolo} & 20  & 73.5 & 60.0 & 54.5 \\ 
RoDLA \cite{rodla} & 342 & \textbf{74.1} & 57.7 & 53.1 \\ 
DINO \cite{dino} & 46 & 69.7 & 53.4 & 49.2 \\ 
DIT \cite{dit} & 86 & 67.2 & 51.8 & 47.8 \\ 
Florence-2 \cite{florence2} & 826 & 41.4 & 28.7 & 28.0 \\ 
\bottomrule
\end{tabular}
\vspace{5pt} 
\caption{Performance comparison on \indicdlp.}
\label{tab:performance_comparison}
\end{table*}

We observe consistent patterns across all evaluated models. For instance, regions requiring semantic understanding, such as \textit{jumpline} (which indicates the continuation of text on another page or section) or \textit{author}, tend to underperform when models rely solely on visual features. Similarly, regions like \textit{index} (list of references at the back of a book) or \textit{subsub-unordered-list} (the third hierarchical level within an unordered list) also perform poorly due to their relatively low frequency in the dataset. In contrast, regions like \textit{placeholder-text} (text entry fields) and \textit{page-number}, though more frequent, show lower performance due to their diverse visual appearances. On the other hand, regions like \textit{footnote} and \textit{flag} (the nameplate/branding of a newspaper/magazine) demonstrate higher accuracy despite their lower frequency, likely due to their consistent appearance and fixed positions, usually at the bottom or top of the document page, respectively.

\begin{figure}[!t]
\centering
\includegraphics[width=0.97\linewidth]{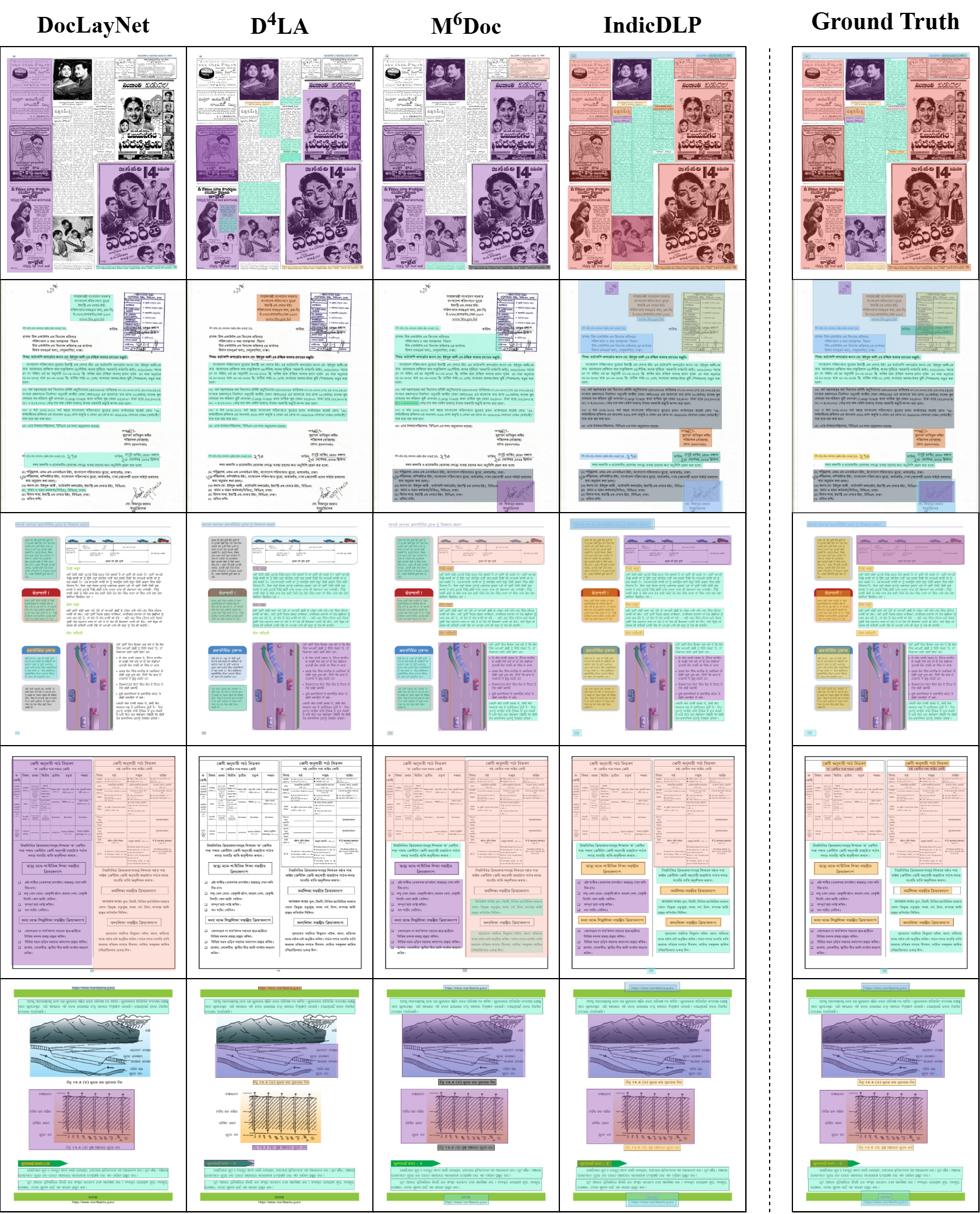}
\caption{Qualitative analysis of YOLOv10x model predictions on \indicdlp~test set when trained from scratch on DocLayNet (Column 1), \dfourla~(Column 2), \msixdoc~(Column 3), and \indicdlp~(Column 4), compared to the Ground Truth (Column 5). For instance, in Rows 1 and 3, for \msixdoc, despite having labels like \textit{advertisements} and \textit{sidebars}, distribution differences prevent accurate localization and classification. In Rows 2 and 4, models trained on \indicdlp~show superior performance on non-standard and multicolumn layouts, respectively. Row 5 highlights partial \textit{figure} detections for DocLayNet and \dfourla, even though it is a common region across all the above datasets.}

\label{fig:comparisons-qualitative}
\end{figure}

On \indicdlp, YOLOv10x~\cite{yolov10} and DocLayout-YOLO~\cite{doclayout-yolo} perform similarly, likely because DocLayout-YOLO~\cite{doclayout-yolo} is a variant of the former with document-specific optimizations in its design. RoDLA~\cite{rodla}, intended to be a more robust version of DINO~\cite{dino}, outperforms DINO, indicating its effectiveness on naturally degraded document images in our dataset. Despite its significantly larger capacity, Florence-2~\cite{florence2} underperforms compared to specialized object detection models, likely due to its pretraining emphasis on diverse natural images rather than document-specific layouts. For further qualitative insights, we present sample predictions from the YOLOv10x model, trained from scratch on DocLayNet, \dfourla, \msixdoc, and \indicdlp. These predictions are compared against the Ground Truth annotations in \Cref{fig:comparisons-qualitative}.

\subsection{How do models perform across the different domains in \indicdlp?}
\label{sec:domains}

\begin{figure*}[!t] 
\centering
\includegraphics[width=\linewidth]{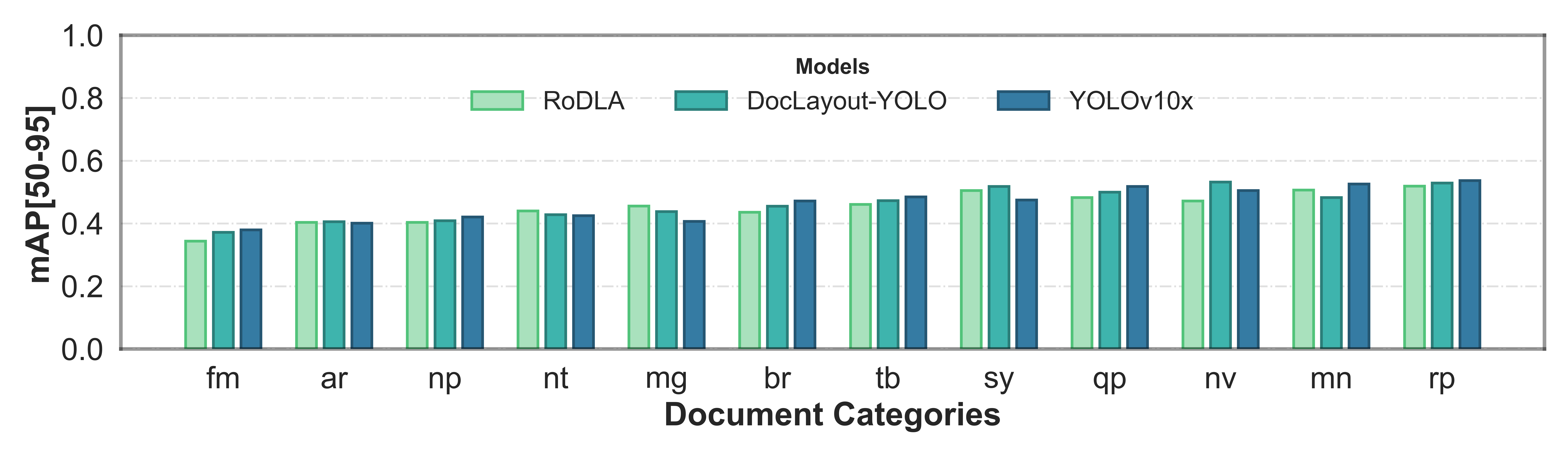}
\caption{Comparison of mAP Scores Across Different Domains for YOLOv10x, DocLayout-YOLO and RoDLA trained on \indicdlp.}
\label{fig:domain-perf}
\end{figure*}

The diversity of document styles across domains in \indicdlp, as illustrated in \Cref{fig:dataset}, leads to substantial variation in domain-wise performance. We show the domain-wise performance of the top 3 best performing models on \indicdlp~in \Cref{fig:domain-perf}. We note that domains such as \textit{Forms} and \textit{Acts \& Rules} have significantly different layouts compared to others, negatively impacting the performance of the model. Certain domains like \textit{Newspapers} and \textit{Brochures} contain a wider range of unique regions to identify, making precise region recognition more complex. Additionally, the dense, multi-column layouts found in \textit{Newspapers} and \textit{Magazines} also make detection tasks more challenging. In contrast, categories with simpler, more uniform, and predictable layouts, such as \textit{Novels} and \textit{Research Papers}, achieve higher mAP scores due to better detection accuracy.

\subsection{How do models perform across the different languages in \indicdlp?}
\label{sec:languages}

\begin{table*}[!t]
\renewcommand{\arraystretch}{1.2}
\scriptsize
\centering
\begin{tabular}{l|*{12}{>{\centering\arraybackslash}p{0.06\textwidth}}}
\toprule
\diagbox[width=2.0cm, height=0.6cm, innerleftsep=1pt, innerrightsep=0.5pt]{\shortstack{\textbf{↓ Train}}}{\shortstack{\textbf{Test} \\ \textbf{→}}} & 
\multicolumn{1}{c}{\textbf{en}} & 
\multicolumn{1}{c}{\textbf{\textcolor{Maroon}{hi}}} & 
\multicolumn{1}{c}{\textbf{\textcolor{Maroon}{mr}}} & 
\multicolumn{1}{c}{\textbf{gu}} & 
\multicolumn{1}{c}{\textbf{pa}} & 
\multicolumn{1}{c}{\textbf{\textcolor{RoyalBlue}{bn}}} & 
\multicolumn{1}{c}{\textbf{\textcolor{RoyalBlue}{as}}} & 
\multicolumn{1}{c}{\textbf{or}} & 
\multicolumn{1}{c}{\textbf{ta}} & 
\multicolumn{1}{c}{\textbf{te}} & 
\multicolumn{1}{c}{\textbf{kn}} & 
\multicolumn{1}{c}{\textbf{ml}} \\
\midrule
\textbf{English   }                                & \cellcolor[HTML]{A3DABF}0.39          & \cellcolor[HTML]{E8F6EF}0.18          & \cellcolor[HTML]{E8F6EF}0.18          & \cellcolor[HTML]{FAFDFC}0.12          & \cellcolor[HTML]{E9F6F0}0.18          & \cellcolor[HTML]{ECF8F2}0.17          & \cellcolor[HTML]{F9FDFB}0.13          & \cellcolor[HTML]{F3FAF7}0.15          & \cellcolor[HTML]{E7F6EF}0.18          & \cellcolor[HTML]{F3FAF7}0.15          & \cellcolor[HTML]{EEF8F3}0.16          & \cellcolor[HTML]{E6F5EE}0.19          \\
\textbf{\textcolor{Maroon}{Hindi}}                                   & \cellcolor[HTML]{E9F6F0}0.18          & \cellcolor[HTML]{90D2B2}0.45          & \cellcolor[HTML]{C9EADA}0.27          & \cellcolor[HTML]{E5F5ED}0.19          & \cellcolor[HTML]{E0F3EA}0.20          & \cellcolor[HTML]{D7EFE3}0.23          & \cellcolor[HTML]{D9F0E5}0.23          & \cellcolor[HTML]{E4F4EC}0.19          & \cellcolor[HTML]{E9F7F0}0.18          & \cellcolor[HTML]{E5F5ED}0.19          & \cellcolor[HTML]{DFF2E9}0.21          & \cellcolor[HTML]{E3F4EB}0.20          \\
\textbf{\textcolor{Maroon}{Marathi}}                         & \cellcolor[HTML]{EDF8F2}0.16          & \cellcolor[HTML]{C8E9D9}0.28          & \cellcolor[HTML]{72C69D}0.55          & \cellcolor[HTML]{E5F5ED}0.19          & \cellcolor[HTML]{EBF7F1}0.17          & \cellcolor[HTML]{D8EFE4}0.23          & \cellcolor[HTML]{D6EFE3}0.24          & \cellcolor[HTML]{E1F3EA}0.20          & \cellcolor[HTML]{E5F5ED}0.19          & \cellcolor[HTML]{F6FCF9}0.14          & \cellcolor[HTML]{E3F4EC}0.20          & \cellcolor[HTML]{E6F5EE}0.18          \\
\textbf{Gujarati}                                   & \cellcolor[HTML]{EEF8F3}0.16          & \cellcolor[HTML]{E2F4EB}0.20          & \cellcolor[HTML]{D9F0E4}0.23          & \cellcolor[HTML]{9BD7B9}0.42          & \cellcolor[HTML]{EAF7F0}0.17          & \cellcolor[HTML]{E0F3E9}0.20          & \cellcolor[HTML]{DCF1E7}0.22          & \cellcolor[HTML]{E6F5EE}0.19          & \cellcolor[HTML]{E9F6F0}0.18          & \cellcolor[HTML]{F4FBF7}0.14          & \cellcolor[HTML]{E9F7F0}0.18          & \cellcolor[HTML]{EDF8F2}0.16          \\
\textbf{Punjabi}                                   & \cellcolor[HTML]{F8FDFA}0.13          & \cellcolor[HTML]{E0F3EA}0.20          & \cellcolor[HTML]{DEF2E8}0.21          & \cellcolor[HTML]{E8F6EF}0.18          & \cellcolor[HTML]{84CDA9}0.49          & \cellcolor[HTML]{E0F3EA}0.20          & \cellcolor[HTML]{E0F3E9}0.21          & \cellcolor[HTML]{E7F6EF}0.18          & \cellcolor[HTML]{F0F9F5}0.15          & \cellcolor[HTML]{F8FCFA}0.13          & \cellcolor[HTML]{EFF9F4}0.16          & \cellcolor[HTML]{EEF9F4}0.16          \\
\textbf{\textcolor{RoyalBlue}{Bengali}}                                   & \cellcolor[HTML]{F3FBF7}0.14          & \cellcolor[HTML]{DFF2E9}0.21          & \cellcolor[HTML]{D7EFE3}0.23          & \cellcolor[HTML]{EEF8F3}0.16          & \cellcolor[HTML]{EAF7F1}0.17          & \cellcolor[HTML]{8FD2B1}0.46          & \cellcolor[HTML]{CFECDE}0.26          & \cellcolor[HTML]{DAF0E6}0.22          & \cellcolor[HTML]{E8F6EF}0.18          & \cellcolor[HTML]{EFF9F4}0.16          & \cellcolor[HTML]{E8F6EF}0.18          & \cellcolor[HTML]{EAF7F1}0.17          \\
\textbf{\textcolor{RoyalBlue}{Assamese}}                                   & \cellcolor[HTML]{F8FCFA}0.13          & \cellcolor[HTML]{D7EFE3}0.23          & \cellcolor[HTML]{D3EEE1}0.24          & \cellcolor[HTML]{F5FBF8}0.14          & \cellcolor[HTML]{E2F4EB}0.20          & \cellcolor[HTML]{CCEBDC}0.27          & \cellcolor[HTML]{70C59C}0.55          & \cellcolor[HTML]{D8F0E4}0.23          & \cellcolor[HTML]{E7F5EE}0.18          & \cellcolor[HTML]{F5FBF8}0.14          & \cellcolor[HTML]{EDF8F2}0.17          & \cellcolor[HTML]{E4F4EC}0.19          \\
\textbf{Odiya}                                   & \cellcolor[HTML]{FCFEFD}0.12          & \cellcolor[HTML]{E8F6EF}0.18          & \cellcolor[HTML]{E2F3EB}0.20          & \cellcolor[HTML]{F4FBF8}0.14          & \cellcolor[HTML]{F1F9F5}0.15          & \cellcolor[HTML]{E1F3EA}0.20          & \cellcolor[HTML]{E0F3EA}0.20          & \cellcolor[HTML]{84CEAA}0.49          & \cellcolor[HTML]{EFF9F4}0.16          & \cellcolor[HTML]{F1FAF5}0.15          & \cellcolor[HTML]{EAF7F1}0.17          & \cellcolor[HTML]{EAF7F0}0.17          \\
\textbf{Tamil}                                   & \cellcolor[HTML]{F5FBF8}0.14          & \cellcolor[HTML]{EAF7F0}0.17          & \cellcolor[HTML]{E9F6F0}0.18          & \cellcolor[HTML]{FBFEFC}0.12          & \cellcolor[HTML]{F0F9F5}0.15          & \cellcolor[HTML]{E9F6F0}0.18          & \cellcolor[HTML]{ECF8F2}0.17          & \cellcolor[HTML]{E7F5EE}0.18          & \cellcolor[HTML]{84CEAA}0.49          & \cellcolor[HTML]{E9F6F0}0.18          & \cellcolor[HTML]{DFF2E9}0.21          & \cellcolor[HTML]{E1F3EA}0.20          \\
\textbf{Telugu}                                   & \cellcolor[HTML]{FEFFFE}0.11          & \cellcolor[HTML]{F4FBF7}0.14          & \cellcolor[HTML]{F6FCF9}0.14          & \cellcolor[HTML]{FFFFFF}0.11          & \cellcolor[HTML]{F5FBF8}0.14          & \cellcolor[HTML]{EEF8F3}0.16          & \cellcolor[HTML]{F4FBF8}0.14          & \cellcolor[HTML]{F4FBF8}0.14          & \cellcolor[HTML]{EEF8F3}0.16          & \cellcolor[HTML]{99D6B8}0.43          & \cellcolor[HTML]{E7F6EF}0.18          & \cellcolor[HTML]{E9F6F0}0.18          \\
\textbf{Kannada}                                   & \cellcolor[HTML]{E9F7F0}0.18          & \cellcolor[HTML]{DCF1E7}0.22          & \cellcolor[HTML]{DDF2E7}0.21          & \cellcolor[HTML]{F9FDFB}0.13          & \cellcolor[HTML]{EFF9F4}0.16          & \cellcolor[HTML]{E7F5EE}0.18          & \cellcolor[HTML]{EEF8F3}0.16          & \cellcolor[HTML]{E9F6EF}0.18          & \cellcolor[HTML]{DBF1E6}0.22          & \cellcolor[HTML]{E2F4EB}0.20          & \cellcolor[HTML]{94D4B5}0.44          & \cellcolor[HTML]{E2F4EB}0.20          \\
\textbf{Malayalam}                                   & \cellcolor[HTML]{E6F5EE}0.18          & \cellcolor[HTML]{F2FAF6}0.15          & \cellcolor[HTML]{E5F5ED}0.19          & \cellcolor[HTML]{FCFEFD}0.12          & \cellcolor[HTML]{F2FAF6}0.15          & \cellcolor[HTML]{EBF7F1}0.17          & \cellcolor[HTML]{F0F9F5}0.16          & \cellcolor[HTML]{E4F5ED}0.19          & \cellcolor[HTML]{DDF1E7}0.21          & \cellcolor[HTML]{EDF8F3}0.16          & \cellcolor[HTML]{DDF2E8}0.21          & \cellcolor[HTML]{89D0AD}0.47          \\
\midrule
\tiny{\textbf{\indicdlp-mini}}                        & \cellcolor[HTML]{BAE3CF}0.32          & \cellcolor[HTML]{ABDDC5}0.37          & \cellcolor[HTML]{8FD2B1}0.45          & \cellcolor[HTML]{BAE3CF}0.32          & \cellcolor[HTML]{ADDEC6}0.36          & \cellcolor[HTML]{A8DCC3}0.38          & \cellcolor[HTML]{8DD1B0}0.46          & \cellcolor[HTML]{A1D9BE}0.40          & \cellcolor[HTML]{9ED8BC}0.41          & \cellcolor[HTML]{B9E3CE}0.32          & \cellcolor[HTML]{B5E1CC}0.34          & \cellcolor[HTML]{9BD7BA}0.42          \\
\textbf{\tiny{\indicdlp}}                             & \cellcolor[HTML]{7FCCA6}\textbf{0.50} & \cellcolor[HTML]{79C9A2}\textbf{0.52} & \cellcolor[HTML]{60BF90}\textbf{0.60} & \cellcolor[HTML]{8AD0AE}\textbf{0.47} & \cellcolor[HTML]{66C194}\textbf{0.58} & \cellcolor[HTML]{76C8A0}\textbf{0.53} & \cellcolor[HTML]{57BB8A}\textbf{0.63} & \cellcolor[HTML]{64C093}\textbf{0.59} & \cellcolor[HTML]{62C092}\textbf{0.60} & \cellcolor[HTML]{84CEAA}\textbf{0.49} & \cellcolor[HTML]{81CCA8}\textbf{0.50} & \cellcolor[HTML]{6FC59B}\textbf{0.55} \\
\bottomrule
\end{tabular}
\vspace{5pt} 
\caption{Zero-shot performance evaluation of YOLOv10x trained on language subsets of \indicdlp. Language codes follow the \textbf{ISO 639-2} standard. Highlighted languages share the same script: \textit{Devanagari} for (\textbf{\textcolor{Maroon}{hi}, \textcolor{Maroon}{mr}}) and \textit{Bengali} for (\textbf{\textcolor{RoyalBlue}{as}, \textcolor{RoyalBlue}{bn}}). The model trained on the full \indicdlp~dataset (last row) significantly outperforms those trained monolingually, benefiting both from a larger data volume and cross-lingual transfer.}

\label{tab:language_comparison}
\end{table*}

Indic scripts across the nation vary significantly in structure and character complexity. The multilingual nature of \indicdlp~offers a valuable opportunity to study how script differences affect layout parsing performance. To support language-specific evaluation, we split \indicdlp~into subsets by language and fine-tune a YOLOv10x model for each. This experimental setup also allows us to assess cross-lingual transfer in layout parsing, particularly between languages with shared scripts or from geographically proximate regions. Specifically, we train on documents in one script and evaluate performance on documents in another.

Quantitative results of this experiment are presented in \Cref{tab:language_comparison}. Comparing the diagonal ‘same script’ entries with those in the last row (i.e. full \indicdlp), we find that a model trained on all scripts outperforms those trained specifically for each script. The substantially larger training set of the full \indicdlp~trained model compensates for the lack of script homogeneity.
To isolate the impact of dataset size, we create a smaller \indicdlp-mini subset, consisting of 10K images proportionally sampled from all languages of \indicdlp~while maintaining consistent domain distribution within each language. The relatively higher performance of the diagonal entries compared to the \indicdlp-mini baseline (second-to-last row in \Cref{tab:language_comparison}) suggests that script influences layout prediction when training data is adequately accounted for. 

From \Cref{tab:language_comparison}, the cross-diagonal results show that zero-shot evaluation on unseen languages generally leads to mAP scores that are 20–25 points lower. This suggests that models trained on a single language rely heavily on script-specific features and generalize poorly to unseen scripts. This, along with the results in the last row, highlights the benefits of incorporating multiple languages in foundational layout datasets. Interestingly, while cross-lingual transfer improves slightly for languages sharing the same script (marked in \textbf{\textcolor{Maroon}{red}} for \textit{Devanagari} and \textbf{\textcolor{RoyalBlue}{blue}} for \textit{Bengali} in \Cref{tab:language_comparison}), non-script-based visual differences (i.e., distribution shift) lead to cross-lingual models for shared-script languages underperforming compared to language-specific models.

\subsection{Does English pretraining improve layout parsing performance?}
\label{english_pt}

\begin{table*}[!t]
\renewcommand{\arraystretch}{1.2}
\centering
{\scriptsize
\begin{tabular}{lcccl}
\toprule
\textbf{Pretraining Dataset(PD)} & 
\textbf{PD Size} & 
\textbf{mAP50} &
\textbf{mAP75} & 
\makebox[0pt][l]{\textbf{mAP[50:95]}} \\
\midrule
\textsc{UED-mini} & 75K  & 76.4 & 63.7 & \textbf{57.7} (\textbf{\textcolor{ForestGreen}{+1.9}}) \\
\textit{No Pretraining (Scratch)} & --  & 74.5 & 61.3 & 55.8 (Baseline) \\
COCO & 118K  & 73.5 & 60.6 & 55.0 (\textbf{\textcolor{Maroon}{-0.8}}) \\
\textsc{UED} & 785K  & 72.8 & 59.4 & 53.9 (\textbf{\textcolor{Maroon}{-1.9}}) \\
\textsc{PubLayNet} & 360K  & 71.7 & 59.2 & 53.0 (\textbf{\textcolor{Maroon}{-2.8}}) \\
\bottomrule
\end{tabular}
}
\vspace{5pt}
\caption{Performance of \textsc{YOLOv10x} pretrained on different sources and fine-tuned on \indicdlp. Pretraining on smaller, diverse datasets like UED-mini yields better results than larger generic datasets.}
\label{tab:yolov10x_pretrained}
\end{table*}

Our preliminary experiments indicate that training on high-quality datasets specific to layout parsing can further improve fine-tuning performance on \indicdlp. This strategy is particularly effective due to the considerable variability in quantity, domain, document types, and visual appearances across existing English-dominant layout parsing datasets. In the following section, we explore which data configuration is more effective: (i) using all available datasets, which provides a larger volume of data, or (ii) curating high-quality, manually annotated subsets, which offer greater data diversity.

To test our approach, we evaluated several layout parsing datasets: PubLayNet~\cite{publaynet}, DocBank~\cite{docbank}, DocLayNet~\cite{doclaynet}, \msixdoc~\cite{m6doc}, and \dfourla~\cite{dfourla}, creating two bundles: \textsc{UED} (Unified Existing Datasets) and a curated subset, \textsc{UED-mini} (see \Cref{tab:yolov10x_pretrained}). \textsc{UED-mini} prioritized human-annotated datasets with diverse domains and logical layout elements. PubLayNet and DocBank were excluded due to automated annotations, and \dfourla\ for its grayscale images with limited inter-domain variability. DocLayNet and \msixdoc\ had inconsistently annotated or underrepresented regions, limiting generalization. We refined and merged their labels iteratively, consolidating semantically similar categories (e.g., \textit{table-caption} and \textit{image-caption} as \textit{caption}; \textit{text} in DocLayNet mapped to \textit{paragraph} in \msixdoc). Less significant labels were removed, resulting in a unified taxonomy of 25 labels. This formed the \textsc{UED-mini} dataset, combining the strengths of both the source datasets.

To assess the impact of pretraining, we compared YOLOv10x trained from scratch on \indicdlp\ with variants pretrained on \textsc{UED}, \textsc{UED-mini}, COCO~\cite{coco}, and PubLayNet, followed by fine-tuning on \indicdlp. Results are presented in \Cref{tab:yolov10x_pretrained}.

We observe that pretraining on English datasets does not consistently improve performance. \textsc{UED}, dominated by synthetic data from PubLayNet and DocBank (700K of 785K images), suffers from limited diversity, which impacts performance more than the domain gap introduced by COCO. In contrast, models pretrained on the curated \textsc{UED-mini} set perform better, indicating the necessity of careful selection and curation of existing datasets. These results highlight the value of both domain diversity and high-quality annotations, with \textsc{UED-mini} proving to be the most effective pretraining source for \indicdlp.

\subsection{Conversely, how effective is \indicdlp~as the initial pretraining dataset?}

\begin{figure}[!t]
    \centering
    \includegraphics[width=\linewidth]{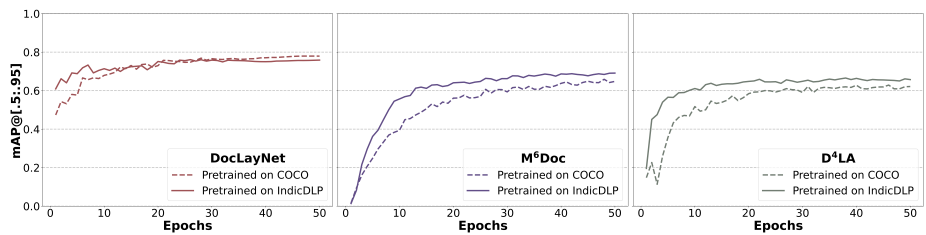}
    \caption{Performance gap observed during finetuning on various datasets using YOLOv10x pretrained on \indicdlp~compared to finetuning from scratch. The solid line represents finetuning with \indicdlp, while the dashed line represents finetuning without it. For \msixdoc~(center) and \dfourla~(right), pretraining on \indicdlp~increases mAP by 2.8 points on average, while for DocLayNet (left), it leads to faster convergence.}
    \label{fig:indicdlp_pretraining}
\end{figure}

In this section, we evaluate the effectiveness of \indicdlp~as a pretraining dataset for layout parsing in other languages. For this, we pretrain two models: one using \indicdlp~and the other using COCO (as the baseline). Both models are then fine-tuned on \msixdoc~\cite{m6doc}, DocLayNet~\cite{doclaynet}, and \dfourla~\cite{dfourla}. 

As shown in \Cref{fig:indicdlp_pretraining}, using \indicdlp~for pretraining offers notable advantages: it either accelerates convergence to optimal values (as observed with DocLayNet \cite{doclaynet}), yields significant performance improvements (as seen with \msixdoc~\cite{m6doc}), or achieves both (in the case of \dfourla~\cite{dfourla}), demonstrating the effectiveness of \indicdlp~as a good pretraining dataset for document layout parsing tasks in other languages.

\section{Conclusion}

Document layout parsing is critical for advancing document understanding, yet existing datasets suffer from limitations in scale, diversity, and annotation granularity. \indicdlp~fills this gap by providing the largest and most diverse dataset for Indic document layout analysis, spanning 12 languages, 12 domains, and 42 region labels. 
We also introduce \textsc{UED-mini} as a curated dataset and demonstrate its utility for pretraining. Our experiments demonstrate that models trained on \indicdlp~generalize well across a wide range of document layouts, improving both Indic and non-Indic document parsing. Our work opens up avenues for enhancing OCR, retrieval, and information extraction for Indic scripts, thereby facilitating inclusive and high-performing document digitization solutions. We release \indicdlp~as an open resource dataset, enabling the community to push the boundaries of multilingual, multi-domain document understanding.

\section{Acknowledgements}

This work is supported by the Ministry of Electronics and Information Technology (MeitY), Government of India, as part of the Digital India Bhashini Mission, which aims to advance Indian language technology. The project’s human resources, including annotators, reviewers, and supercheckers, were supported through a dedicated grant provided to IIT Madras, which serves as the Data Management Unit for the mission.

%
%
\bibliographystyle{splncs04}
\bibliography{main}
\appendix
%
%
\clearpage
\section*{Supplementary Material}
\addcontentsline{toc}{section}{Supplementary Material}

\section{Dataset Comparison}

A key strength of \indicdlp{} is its generalization capability compared to other layout parsing datasets. We provide qualitative evidence of this in the main paper. However, due to space constraints, a detailed quantitative evaluation was not included there; we present it below in \Cref{tab:suppli}. Since different datasets use varying label sets and annotation styles, direct comparisons are not feasible. Therefore, we focus on the common labels shared across datasets.



\begin{table*}[h]
    \scriptsize
    \centering
    \renewcommand{\arraystretch}{1.1} 
    \begin{tabular}{l l c c c c}
        \toprule
        \multicolumn{2}{c}{\textbf{Trained on}} & \multicolumn{4}{c}{\textbf{Testing on}} \\
        \cmidrule(lr){1-2} \cmidrule(lr){3-6}
        & \textbf{Labels} & \textbf{\indicdlp} & \textbf{\msixdoc} & \textbf{\dfourla} & \textbf{DocLayNet} \\
        \midrule
        \multirow{3}{*}{\textbf{\indicdlp}} & Paragraph & 85 & 76 & 59 & 49 \\
        & Table & 85 & 73 & 31 & 67 \\
        & Figure & 78 & 69 & 28 & 43 \\
        \midrule
        \multirow{3}{*}{\textbf{\dfourla}} & Paragraph & 49 & 68 & 93 & 61 \\
        & Table & 46 & 25 & 94 & 20 \\
        & Figure & 38 & 47 & 91 & 24 \\
        \midrule
        \multirow{3}{*}{\textbf{\msixdoc}} & Paragraph & 53 & 92 & 43 & 50 \\
        & Table & 58 & 86 & 12 & 45 \\
        & Figure & 45 & 86 & 15 & 37 \\
        \midrule
        \multirow{3}{*}{\textbf{DocLayNet}} & Paragraph & 41 & 71 & 64 & 75 \\
        & Table & 62 & 54 & 17 & 87 \\
        & Figure & 45 & 57 & 11 & 78 \\
        \bottomrule
    \end{tabular}
    \vspace{5pt} 
    \caption{Mean Average Precision (mAP) [50:95] performance of common region labels for YOLOv10x model trained on one dataset and tested on others. \indicdlp{} demonstrates the best generalization across datasets due to its diverse document layouts, highlighting its applicability to language-agnostic visual document layout parsing.}
    \label{tab:suppli}
\end{table*}


As expected, performance is highest when a model is evaluated on its own test set. However, models trained on datasets other than \indicdlp{} exhibit lower performance on other datasets, despite the fact that the region labels in question are basic physical regions that can typically be distinguished based on visual appearance alone. In contrast, a model trained on \indicdlp{} shows a much smaller performance drop when tested on other datasets while maintaining high accuracy on its own test set. This demonstrates the fact that \indicdlp{}'s diverse document layouts enable strong generalization to distributions that differ significantly, further highlighting its robustness.

\section{Distribution of \indicdlp{} across domains and languages}

\begin{figure*}[] 
\centering
\includegraphics[width=0.8\linewidth]{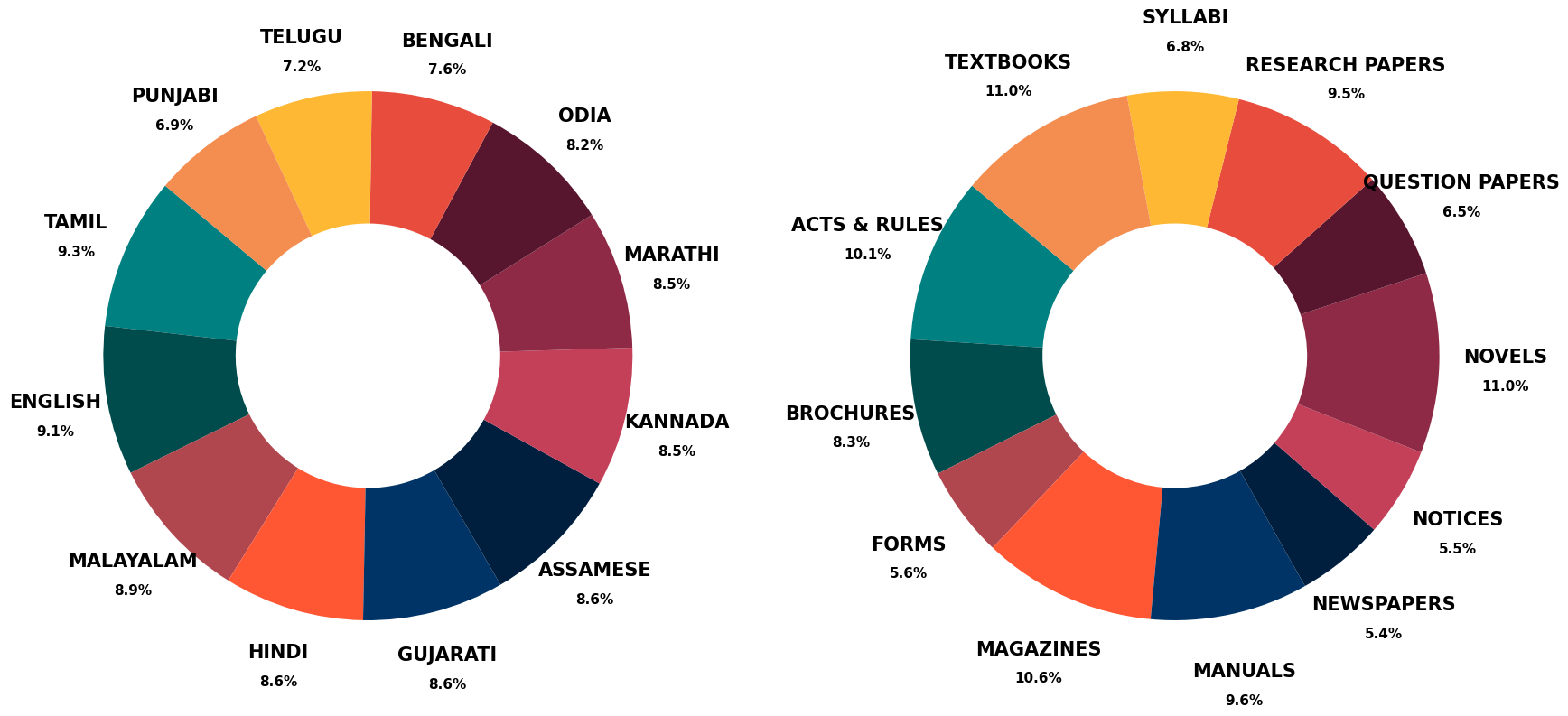}
\caption{The above figure illustrates the contributions of 12 languages (left) and 12 document domains (right) in the \indicdlp{} dataset. The distribution is fairly balanced across both categories, with no single language or domain overwhelmingly dominating the dataset. This ensures a diverse and well-represented collection.}
\label{fig:lang-donut}
\end{figure*}

To ensure that \indicdlp{} supports robust and generalizable document layout detection models, the dataset is designed to capture a diverse range of languages and document domains. As shown in \Cref{fig:lang-donut}, \indicdlp{} includes documents from 11 Indic languages and English, providing broad linguistic representation. This ensures that models trained on our dataset can generalize well across different scripts, many of which feature complex typographical structures, inflections, and diacritical marks that influence layout patterns.

In addition to linguistic diversity, \indicdlp{} spans 12 distinct document domains, covering a wide range of structured and unstructured formats such as newspapers, research papers, textbooks, forms, and novels. Each document type exhibits unique layout structures, font styles, and content organization, all of which impact layout detection. By including such a broad spectrum of document formats, \indicdlp{} ensures that trained models are capable of handling various real-world document structures rather than overfitting to a narrow subset.

From the distribution shown in \Cref{fig:lang-donut}, we show that the contributions of language and domain are relatively balanced, ensuring that no single category dominates overwhelmingly. This equitable distribution enables \indicdlp{} to serve as a fair benchmark for layout analysis across diverse real-world documents. Additionally, the dataset includes documents spanning multiple decades, incorporating typographical variations, print degradation, and non-standard formatting. These real-world complexities present challenges for layout detection but also enhance the dataset’s ability to train models that can effectively handle noisy, degraded, and historically diverse document layouts.

\section{Annotation Interface for \indicdlp{} creation}
All annotations were performed using Shoonya, an open-source annotation framework based on Label Studio. Metadata was meticulously tracked in Shoonya, including annotation performance, average annotation time, draft mode usage for resolving complex or ambiguous cases, and records of documents returned for correction.

\begin{figure*}[] 
\centering
\includegraphics[width=\linewidth]{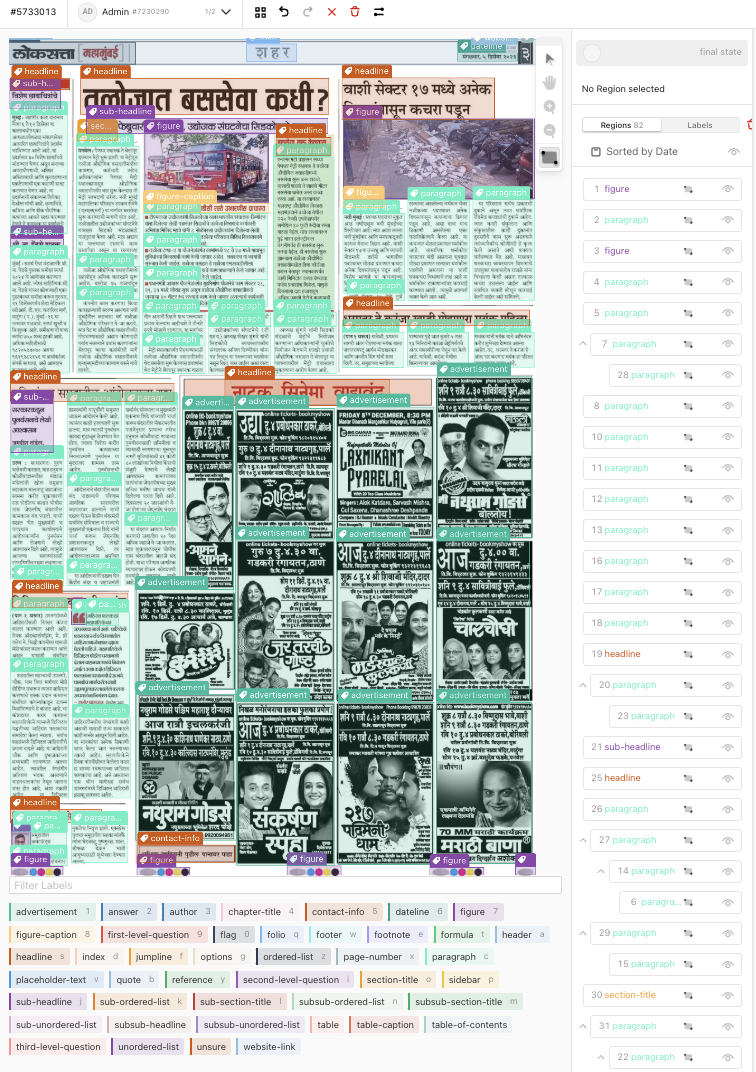}
\caption{Shoonya Annotation Interface}
\label{fig:shoonya_interface}
\end{figure*}

The right sidebar allows users to toggle the region label view using the 'eye' icon. Annotators use the main panel to demarcate and label elements on the page through the label element tabs at the bottom. These tabs are color-coded to enhance visual distinction, particularly when annotating complex documents, as shown in \Cref{fig:shoonya_interface}. 

\section{Annotation and Verification Workflow}
To complete the annotation process within a realistic timeframe, a team of 50 individuals was hired, consisting of 3 to 4 annotators and 1 reviewer per language. All annotators and reviewers were native speakers of the respective language. Additionally, a smaller team of 6 multilingual validators was recruited to supercheck reviewers' work for cross-language inconsistencies.

Reviewers and validators were trained on small, separate subsets of the dataset to ensure accurate annotation and review, even in ambiguous cases. Their insights contributed to the development of a 150-page annotation guideline. To minimize document bias, images were shuffled within languages and domains, ensuring a balanced distribution of complex and simple document layouts among annotators. Each annotated page was thoroughly reviewed for imprecise region boxing and incorrect label assignment. Minor mistakes were corrected immediately, while major errors were sent back to the annotator for revision to reinforce learning. The same process was followed by the validators, ensuring continuous improvement in the team's annotation accuracy. The combined process of annotation, review, and validation took approximately 8 months to complete.

\newpage

\section{Annotated Instances from Guidelines}

Below, we provide annotated examples of regions that are challenging, or less commonly known, as outlined in our guidelines. The goal is to highlight differences between our dataset and existing layout parsing datasets, particularly in how regions should be annotated across varied appearances, positions on the document page, and scripts.

\subsection{\textit{Jumpline}}

A jumpline is a typographic element used in newspapers to indicate the continuation of an article on another page or section. It typically appears at the end of a column, such as "Continued on page 45" or at the top of a column, such as "Continued from page 16".

\begin{figure*}[] 
\centering
\includegraphics[width=\linewidth]{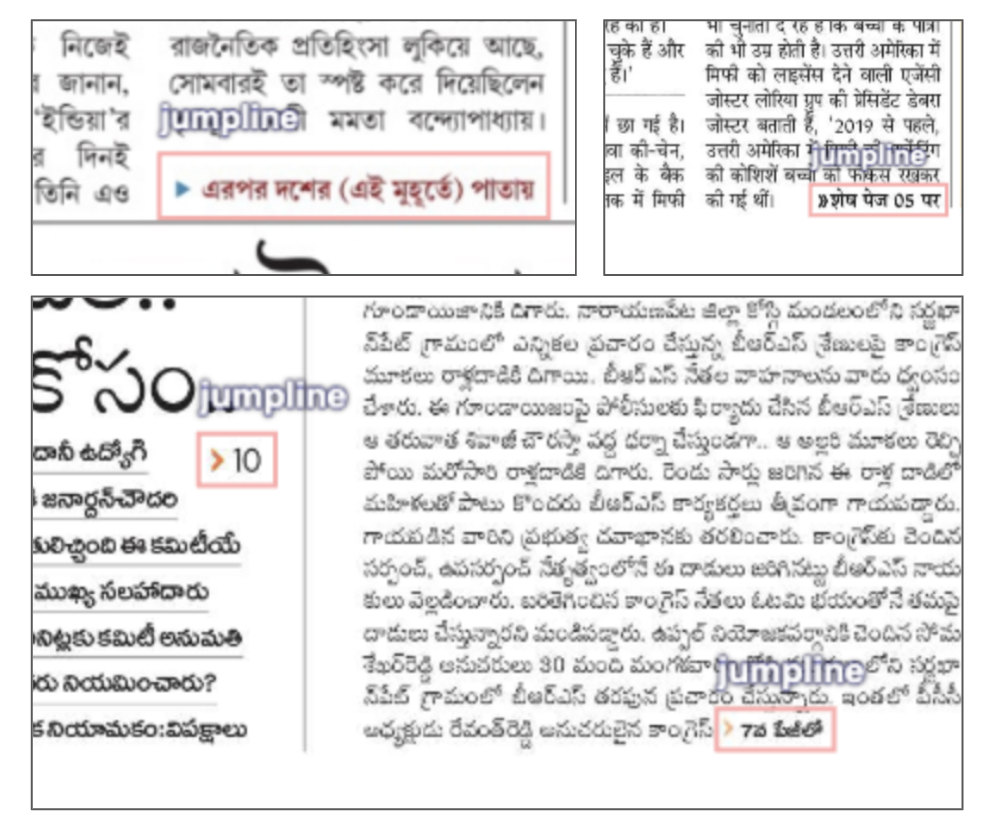}
\caption{Annotated instances of \textit{juumpline} region in \indicdlp.}
\label{fig:domain-perf}
\end{figure*}

\newpage

\subsection{\textit{Flag}}

The flag is a text element located at the top of the front page of newspapers, magazines, and brochures. It typically includes the publication's nameplate and logo and is usually set in the largest font on the page.

\begin{figure*}[] 
\centering
\includegraphics[width=\linewidth]{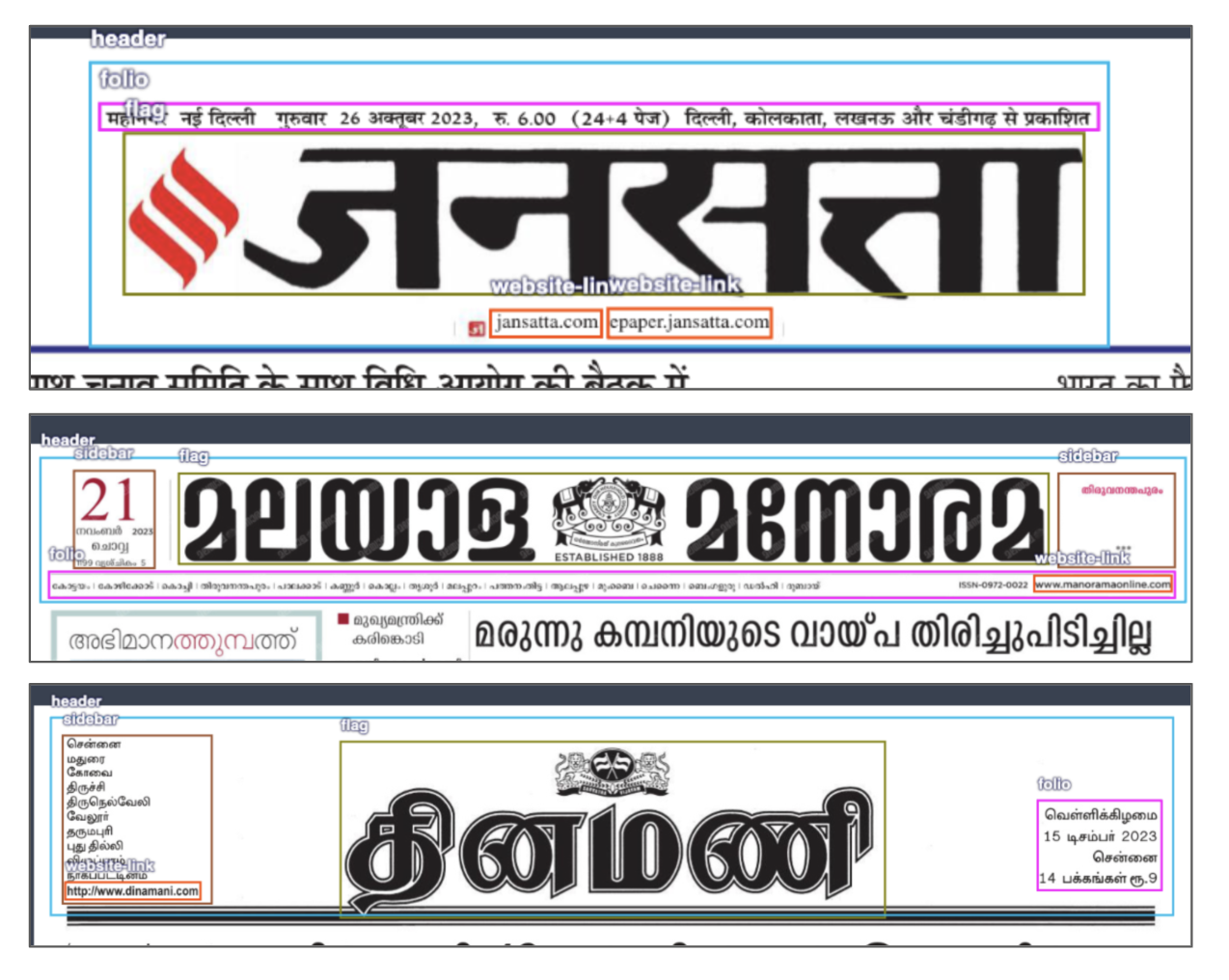}
\caption{Annotated instances of \textit{flag} region in \indicdlp.}
\label{fig:domain-perf}
\end{figure*}
\newpage
\subsection{\textit{Sidebar}}

A sidebar is a short column of text positioned next to the main content and is often visually distinguished by a different background color, text color, or another form of separation. It may also include images, charts, or other visual elements.

\begin{figure*}[] 
\centering
\includegraphics[width=\linewidth]{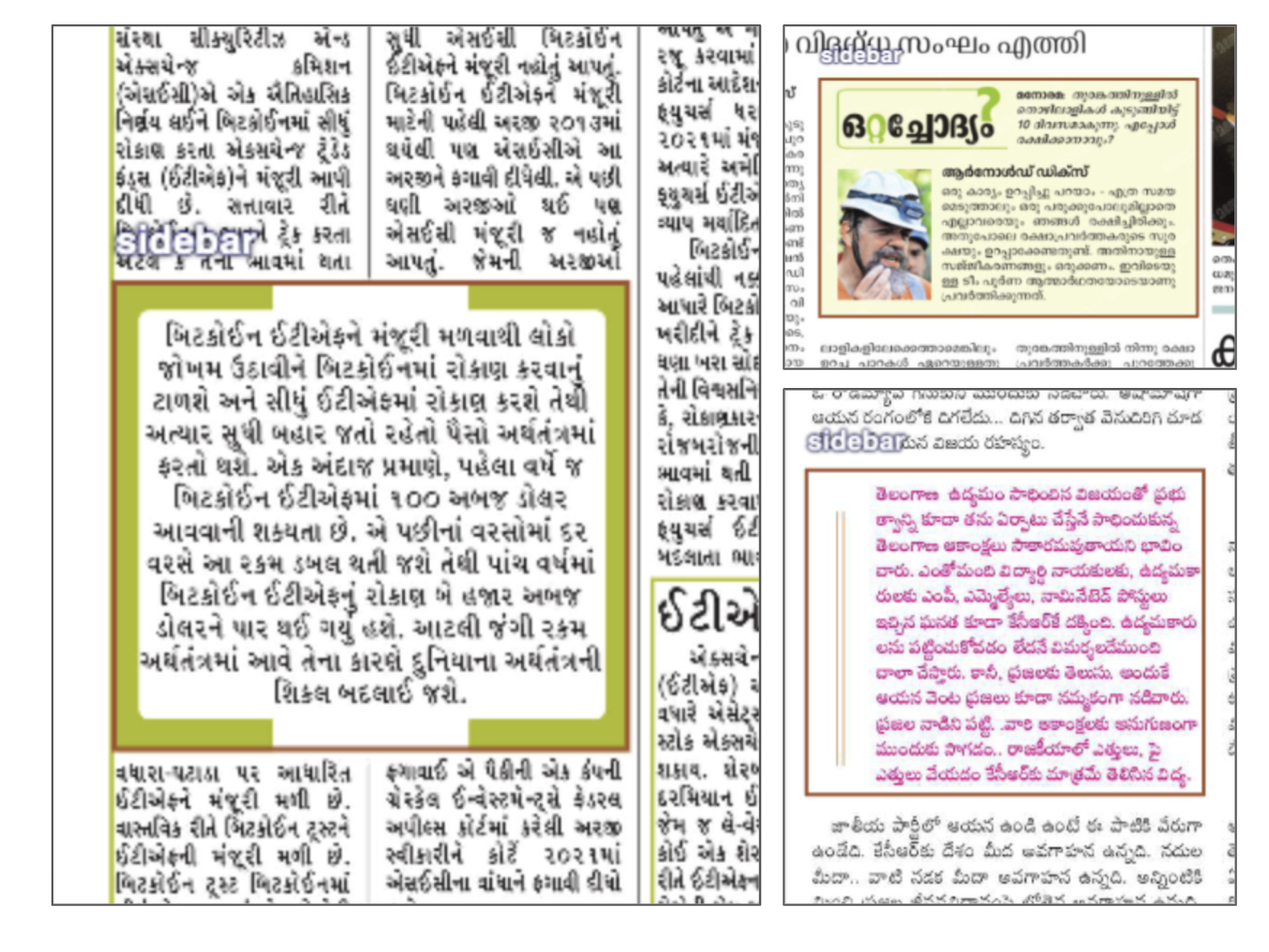}
\caption{Annotated instances of \textit{sidebar} region in \indicdlp.}
\label{fig:domain-perf}
\end{figure*}
\newpage
\subsection{\textit{Header}}
The header is typically located within the top 10–20\% of the page. It may contain text, graphics, a folio, or a page number if present. The header serves as a container for all the elements within this region.

\begin{figure*}[] 
\centering
\includegraphics[width=\linewidth]{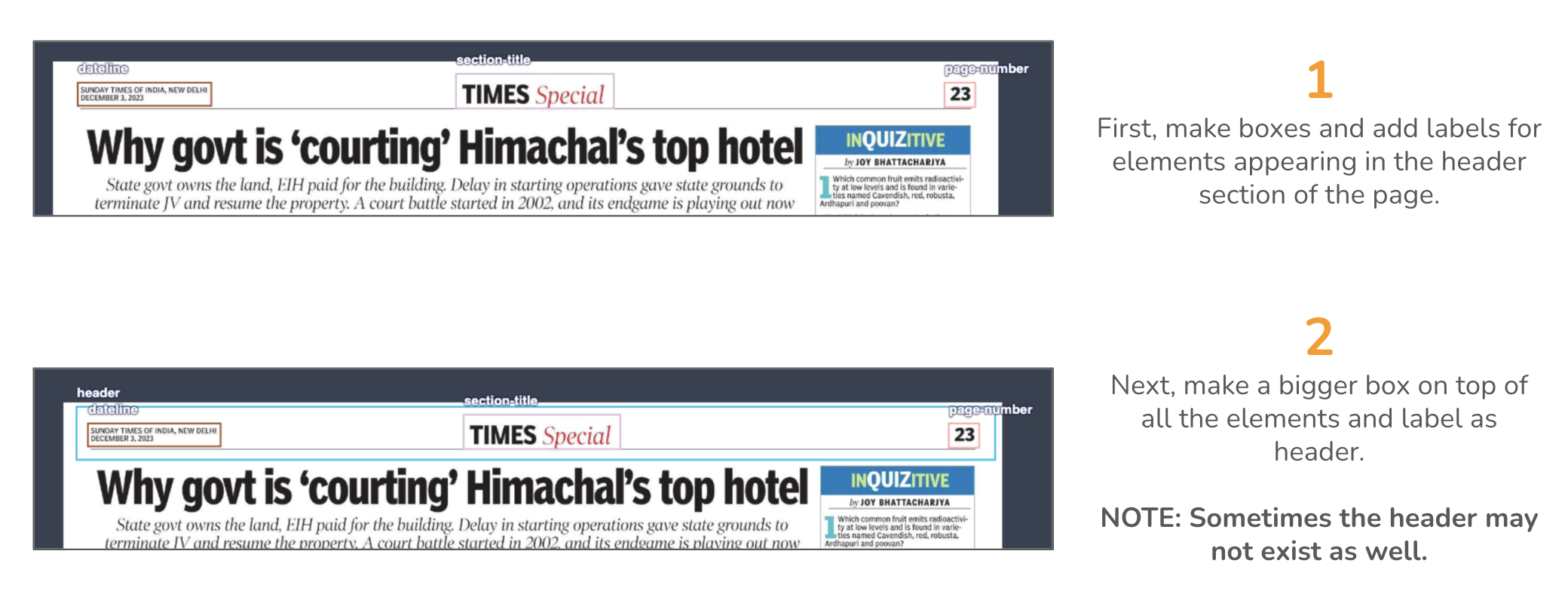}
\caption{Guidelines for annotating a \textit{header} region in \indicdlp.}
\label{fig:domain-perf}
\end{figure*}

\subsection{\textit{Footer}}
The footer is typically located within the bottom 10–20\% of the page. It may contain text, graphics, footnotes, acknowledgments, website links, or page numbers if present. The footer serves as a container for all elements within this region.

\begin{figure*}[] 
\centering
\includegraphics[width=\linewidth]{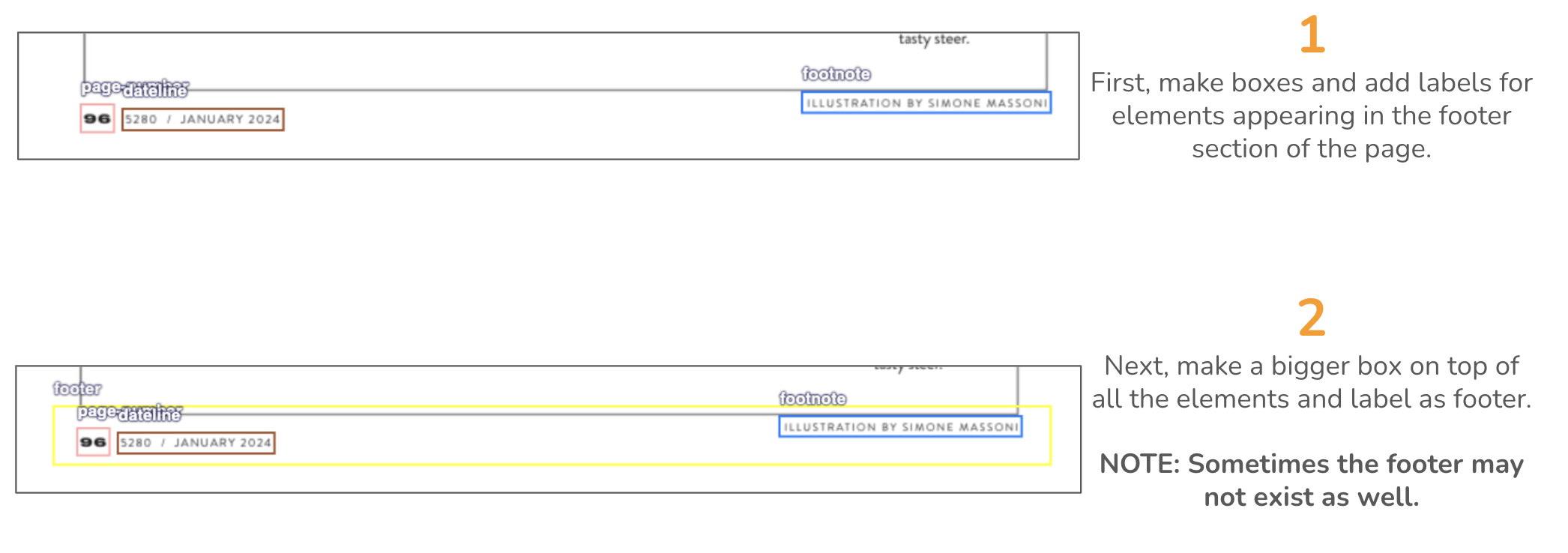}
\caption{Guidelines for annotating a \textit{footer} region in \indicdlp.}
\label{fig:domain-perf}
\end{figure*}

\newpage

\subsection{\textit{Folio}}
A folio is a line of text located in the header or footer of a page. It may include a chapter title (in novels and magazines), a section title (in newspapers), the author's name, the publication date, or other relevant information.

\begin{figure*}[] 
\centering
\includegraphics[width=\linewidth]{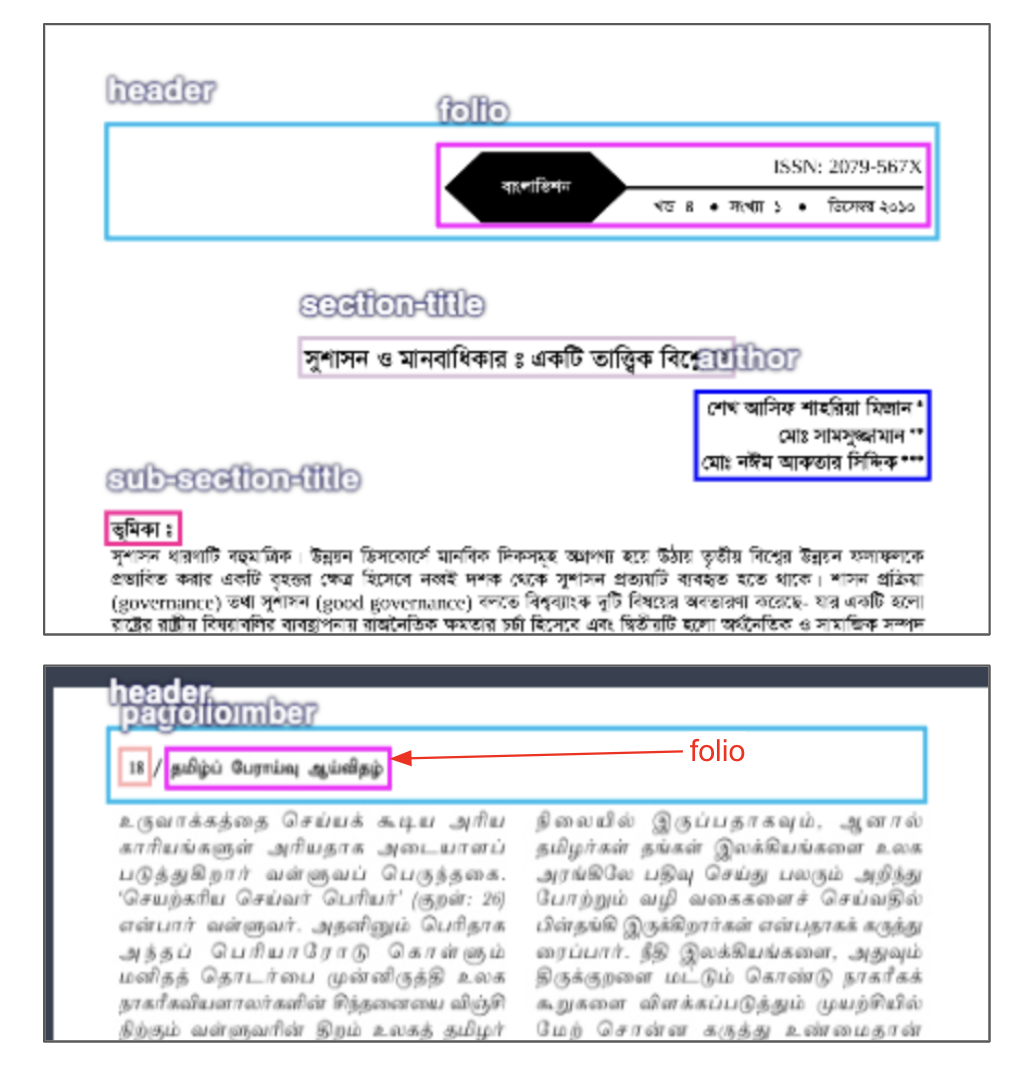}
\caption{Annotated instances of \textit{folio} region in \indicdlp.}
\label{fig:domain-perf}
\end{figure*}
\newpage
\subsection{\textit{Quote}}
In comic book pages, spoken text (blurb) is labeled as a "quote" rather than a "figure caption." This distinction follows the annotation guidelines set by the authors.

\begin{figure*}[] 
\centering
\includegraphics[width=0.8\linewidth]{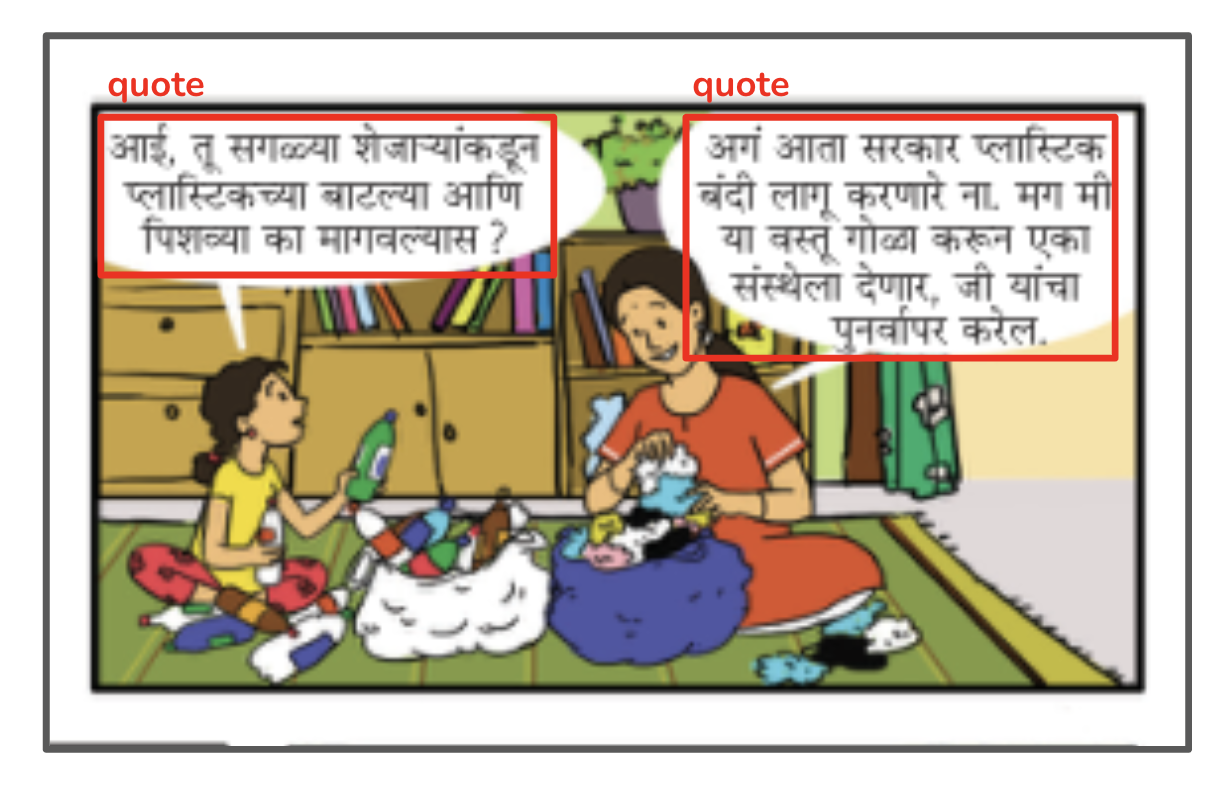}
\caption{Annotated instances of \textit{quote} region in \indicdlp.}
\label{fig:domain-perf}
\end{figure*}

\subsection{\textit{Placeholder-text}}

Placeholder text refers to a blank space, empty rectangle, or fill-in-the-blank area reserved for entering text. It is commonly found in forms but may also appear in textbooks. The key-value pair, consisting of the placeholder box and its corresponding header, is enclosed within this region label box.

\begin{figure*}[] 
\centering
\includegraphics[width=0.7\linewidth]{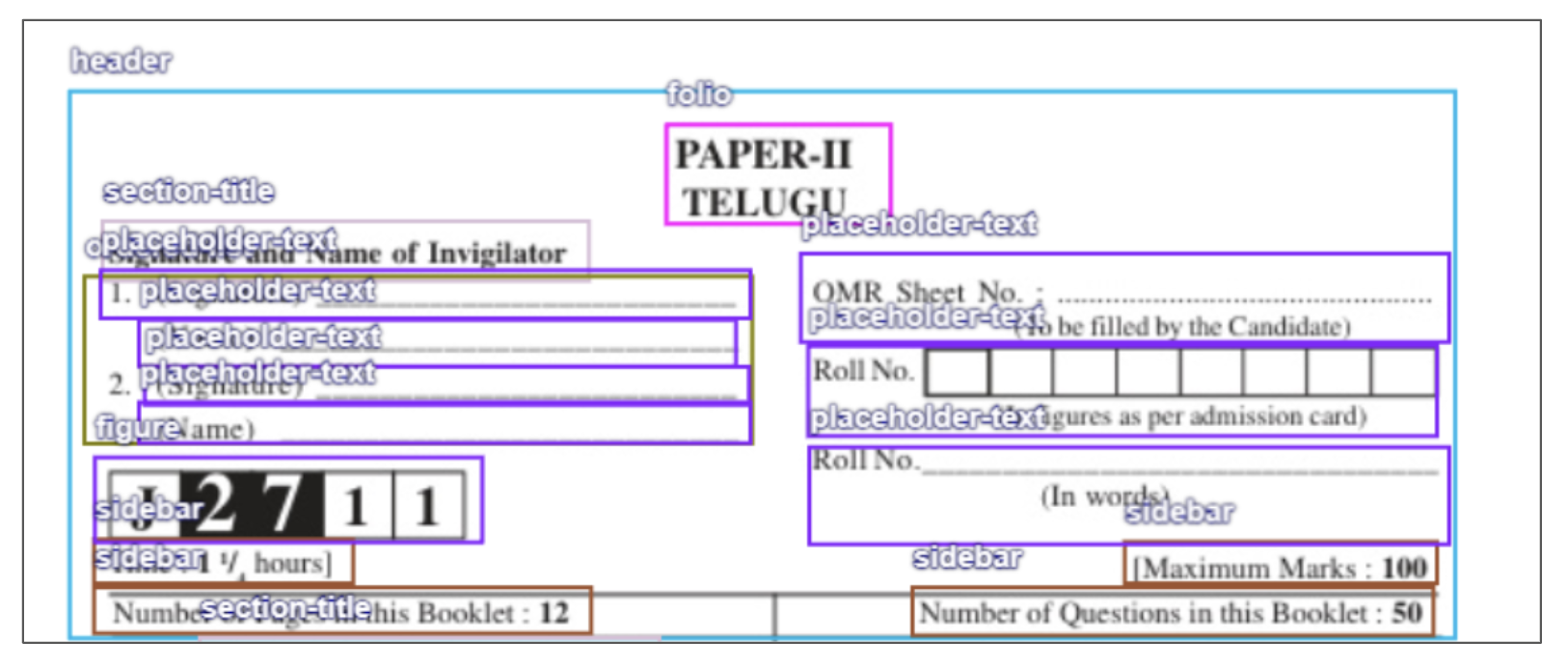}
\caption{Annotated instances of \textit{placeholder-text} region in \indicdlp.}
\label{fig:domain-perf}
\end{figure*}

\newpage

\subsection{\textit{Hierarchical Regions, e.g. Ordered List}}

\indicdlp~includes multiple instances of hierarchical regions, such as \textit{section titles}, \textit{ordered lists}, and \textit{unordered lists}. For these regions, nested and subnested elements are annotated within their respective enclosing region boxes.
\begin{figure*}[] 
\centering
\includegraphics[width=\linewidth]{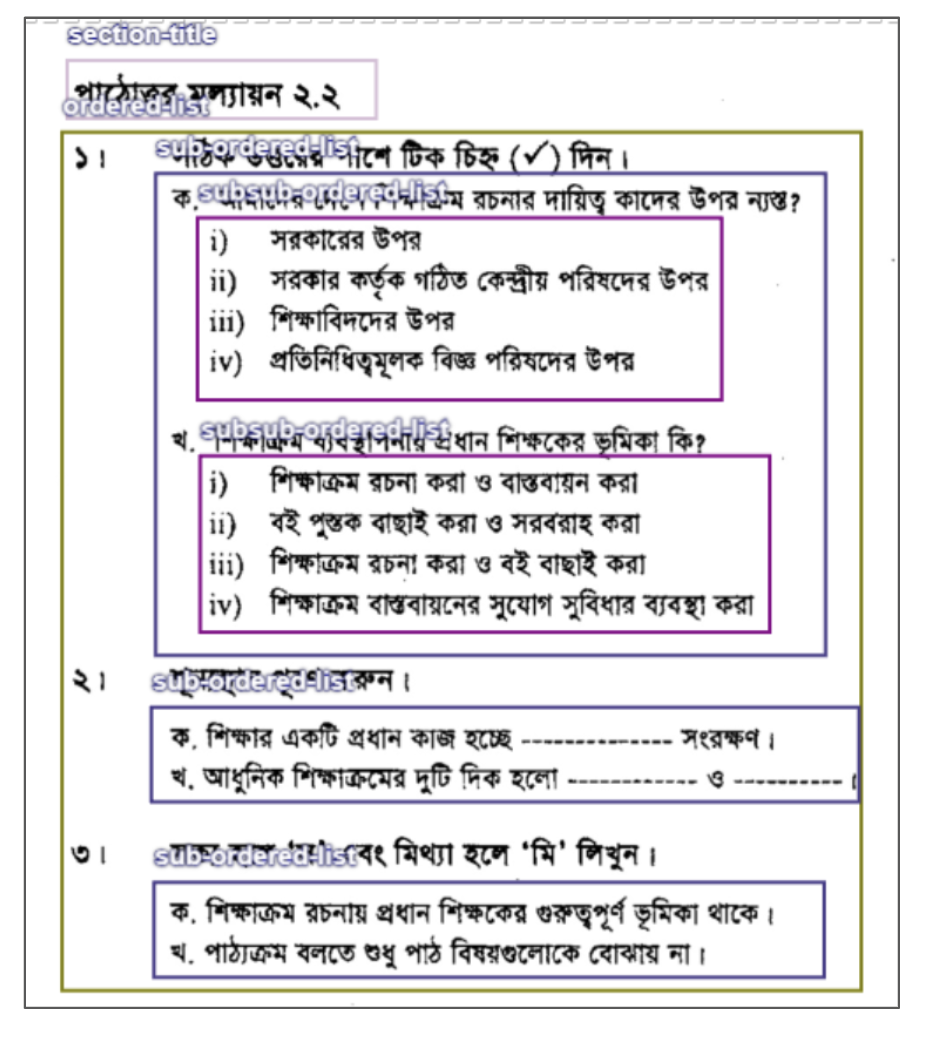}
\caption{Annotated instances of nested and subnested\textit{ordered-list} region in \indicdlp.}
\label{fig:domain-perf}
\end{figure*}



\end{document}